\documentclass[lettersize,journal]{IEEEtran}
\usepackage{subfigure}
\usepackage{graphicx}
\usepackage{amsmath}
\usepackage{amssymb}
\usepackage{booktabs}
\usepackage{algorithmic}
\usepackage{algorithm}
\usepackage{makecell}
\usepackage{bm}
\usepackage{multirow}
\usepackage{dsfont}
\usepackage{ragged2e}
\usepackage[pagebackref,breaklinks,colorlinks]{hyperref}
\usepackage{diagbox}

\begin{document}
\title{HPCR: Holistic Proxy-based Contrastive Replay\\for Online Continual Learning}

\author{Huiwei Lin, Shanshan Feng, Baoquan Zhang, Xutao Li, and Yunming Ye

\thanks{21-Sep-2024
    Manuscript received 26 March 2024; revised 21 September 2024; accepted 31 December 2024. Date of publication XX XXXX 2024; date of current version XX XXXX 2024. This work was supported in part by National Nature Science Foundation of China under Grant 62272130 and Grant 62376072; in part by Nature Science Program of Shenzhen under Grant JCYJ20210324120208022 and Grant JCYJ20200109113014456; in part by Shenzhen Science and Technology Program under Grant KCXFZ20211020163403005. (Corresponding authors: Yunming Ye; Shanshan Feng.)

    Huiwei Lin, Baoquan Zhang, Xutao Li, Yunming Ye are with the Department of Computer Science, Harbin Institute of Technology, Shenzhen 518055, China, and also with Shenzhen Key Laboratory of Internet Information Collaboration, Shenzhen 518055, China (e-mail: yeyunming@hit.edu.cn).

    Shanshan Feng is with the Centre for Frontier AI Research, Institute of High Performance Computing, A*STAR, Singapore (e-mail: victor\_fengss@foxmail.com).
 
 Color versions of one or more figures in this article are available at https://doi.org/XX.XXXX/TNNLS.2024.XXXXXXX.

 Digital Object Identifier XX.XXXX/TNNLS.2024.XXXXXXX
 }
}
 



\markboth{Journal of \LaTeX\ Class Files,~Vol.~14, No.~8, August~2021}%
{Shell \MakeLowercase{\textit{et al.}}: A Sample Article Using IEEEtran.cls for IEEE Journals}

\IEEEpubid{\begin{minipage}{\textwidth}\ \centering Copyright \copyright 2024 IEEE. Personal use of this material is permitted.\\
However, permission to use this material for any other purposes must be obtained from the IEEE by sending an email to pubs-permissions@ieee.org.\end{minipage}}

\maketitle

\begin{abstract}
Online continual learning, aimed at developing a neural network that continuously learns new data from a single pass over an online data stream, generally suffers from catastrophic forgetting. Existing replay-based methods alleviate forgetting by replaying partial old data in a proxy-based or contrastive-based replay manner, each with its own shortcomings. Our previous work proposes a novel replay-based method called proxy-based contrastive replay (PCR), which handles the shortcomings by achieving complementary advantages of both replay manners. In this work, we further conduct gradient and limitation analysis of PCR. The analysis results show that PCR still can be further improved in feature extraction, generalization, and anti-forgetting capabilities of the model. Hence, we develop a more advanced method named holistic proxy-based contrastive replay (HPCR). HPCR consists of three components, each tackling one of the limitations of PCR. The contrastive component conditionally incorporates anchor-to-sample pairs to PCR, improving the feature extraction ability. The second is a temperature component that decouples the temperature coefficient into two parts based on their gradient impacts and sets different values for them to enhance the generalization ability. The third is a distillation component that constrains the learning process with additional loss terms to improve the anti-forgetting ability. Experiments on four datasets consistently demonstrate the superiority of HPCR over various state-of-the-art methods.
\end{abstract}


\ifCLASSOPTIONcompsoc
\IEEEraisesectionheading{\section{Introduction}\label{sec:introduction}}
\else
\section{Introduction}
\label{sec:introduction}
\fi

\IEEEPARstart{O}{nline} continual learning (OCL) is a special scenario of continual learning~\cite{wang2024comprehensive}. Taking the classification problem~\cite{liu2022model} as an example, OCL aims to develop a neural network that can accumulate knowledge of new classes without forgetting information learned from old classes. It is essential to note that the data stream encountered in OCL is non-stationary, with each sample being accessed only once during training. Currently, the main challenge of OCL is catastrophic forgetting (CF)~\cite{kong2023overcoming}, wherein the model experiences a significant decline in performance for old classes when learning new classes\cite{zhao2021memory}.

\begin{figure*}[t]
\centering
\includegraphics[scale=0.52]{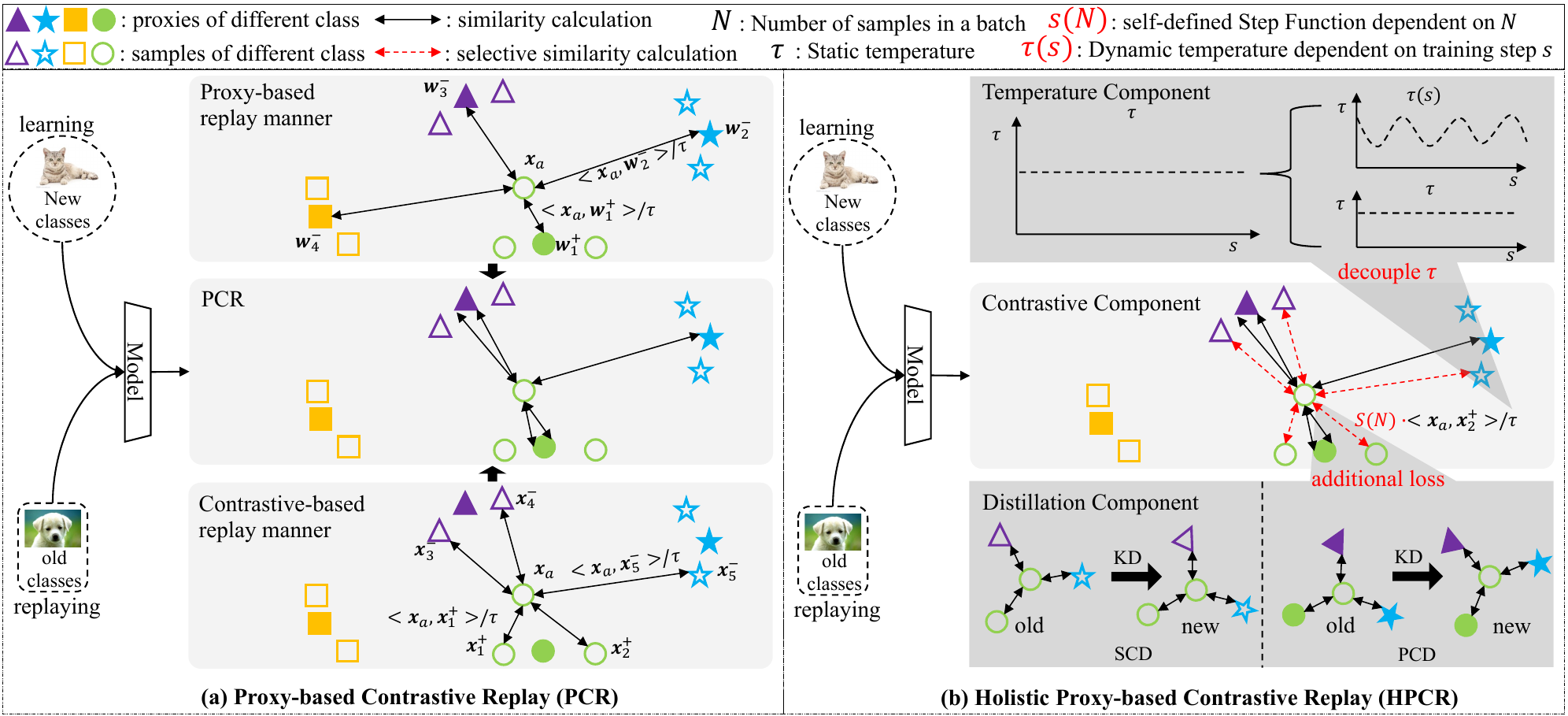}
\vspace{-2ex}
\caption{Illustration of our work. (a) The example of existing replay manners. For each anchor sample $\bm{x}_a$, the proxy-based replay manner calculates similarities of all anchor-to-proxy pairs; the contrastive-based replay manner calculates similarities of all anchor-to-sample pairs in the same training batch; PCR only calculates similarities of selective anchor-to-proxy pairs in the same training batch. (b) The example of the proposed HPCR. The contrastive component conditionally produces anchor-to-sample pairs to PCR; the temperature component decouples the temperature coefficient into two parts based on their impacts on gradients and sets them differently; the distillation component constrains new anchor-to-sample pairs using the old ones (SCD), and distills old anchor-to-proxy pairs to the new ones (PCD).}
\label{fig:illustration}%
\vspace{-2ex}
\end{figure*}

The replay-based methods~\cite{lin2022anchor}, which replay a portion of previous samples, have shown superior performance for OCL~\cite{mai2022online}. In general, there are two ways to replay as shown in Fig.~\ref{fig:illustration}(a). The first is the proxy-based replay manner, which replays using a proxy-based loss function and softmax classifier. It calculates similarities between each anchor and the proxies belonging to all classes. Here, a proxy can be regarded as the representative of a class~\cite{yao2022pcl}, and the anchor represents a sample in a training batch. This manner is subjected to the ``bias'' issue resulting from class imbalance~\cite{lin2022anchor}. The second is the contrastive-based replay manner, which replays using a contrastive-based loss function and nearest class mean (NCM) classifier~\cite{mensink2013distance}. It computes similarities between each anchor and all samples in the same training batch. This manner is unstable and hard to converge during training. To sum up, these two manners have their corresponding shortcomings.  

\IEEEpubidadjcol

Our previous work~\cite{Lin_2023_CVPR} proposes proxy-based contrastive replay (PCR) method to handle these shortcomings. Specifically, we perform a comprehensive analysis and find that coupling both replay manners achieves complementary advantages. The proxy-based replay manner enables fast and reliable convergence with the help of proxies, thereby overcoming the problems of being unstable and hard to converge in the contrastive-based replay manner. Meanwhile, the contrastive-based replay manner can provide the proxy-based replay manner with a new way of selecting anchor-to-proxy pairs, which has been proven to effectively address the ``bias'' issue~\cite{ahn2021ss,caccia2021new}. With these inspirations, PCR calculates similarities between each anchor and the proxies whose associated classes of samples appear in the same batch as illustrated in Fig.~\ref{fig:illustration}(a). 

In this work, we conduct an in-depth analysis of PCR and explore its potential for improvement. Specifically, we derive the gradient propagation process of PCR in the classifier based on the one for general proxy-based loss. The gradient analysis shows that PCR benefits from the number of samples of different classes in each training batch, further confirming its effectiveness and reliability. However, our limitation analysis reveals that PCR still has shortcomings in three key areas: 1) PCR considers the relations between anchor-to-proxy pairs but ignores the relations between anchor-to-sample pairs. It results in insufficient \textbf{feature extraction capability} of the model, particularly in scenarios with large training batches where anchor-to-sample pairs contain important semantic information. 2) The simplistic setting of the temperature coefficient in PCR overlooks its impact on the gradient propagation process. This oversight undermines the model's \textbf{generalization capability}, as the use of a static constant fails to optimize effectively. 3) Old classes included in the learning process alongside new classes continue to encounter bias issues during training. It implies that there is still room for improvement in the model's \textbf{anti-forgetting capability}. Hence, PCR can be enhanced by overcoming these limitations.

To this end, we develop a more comprehensive method named holistic proxy-based contrastive replay (HPCR). As illustrated in Fig.~\ref{fig:illustration}(b), HPCR consists of three components. 1) The \textbf{contrastive component} incorporates anchor-to-sample pairs conditionally into PCR to improve the model's feature extraction capability, capturing more fine-grained semantic information when dealing with large training batches. Specifically, we introduce a step function $s(N)$ to regulate the importance of the anchor-to-sample pairs. When the number of samples is small, their contribution is set as 0; otherwise, it is 1. 2) Meanwhile, the \textbf{temperature component} decomposes the temperature coefficient into two parts based on their influence on gradients. One part is assigned a static constant value, while the other is denoted as a dynamic function value to strengthen the generalization capability, promoting the acquisition of novel knowledge. 3) Additionally, the \textbf{distillation component} imposes further constraints on the replaying process to enhance the anti-forgetting capability, preserving more historical knowledge. This component consists of two knowledge distillation (KD) mechanisms: proxy-based contrastive distillation (PCD) and sample-based contrastive distillation (SCD). The former distills the old anchor-to-proxy correlations into the new ones, while the latter constrains the new anchor-to-sample correlations based on the old ones. 

Our main contributions can be summarized as follows:
\begin{itemize}
\item[1)] We perform further analyses of the PCR. The gradient analysis verifies the effectiveness and reliability of PCR, while the limitation analysis suggests that PCR still has three limitations and can be further improved.

\item[2)] We develop a more holistic method based on PCR, named HPCR, which consists of a contrastive component, a temperature component, and a distillation component. These three components improve the model's feature extraction, generalization, and anti-forgetting capabilities, respectively.

\item[3)] We conduct extensive experiments on four datasets, and the empirical results consistently demonstrate the superiority of HPCR over various state-of-the-art methods. We also investigate and analyze the benefits of each component by ablation studies. And the codes are open-sourced at \url{https://github.com/FelixHuiweiLin/PCR}.
\end{itemize}

This work is an extension of our conference version presented in~\cite{Lin_2023_CVPR}. It offers a more comprehensive approach, which can be delineated into four aspects: (i) This version includes a gradient analysis of PCR, which enhances its theoretical foundation and helps us discover its limitations. (ii) This version conditionally integrates anchor-to-sample pairs into PCR to augment the model's feature extraction capability, treating it as a contrastive component. (iii) This version introduces a temperature component to enhance the model's generalization ability and a distillation component to improve the model's anti-forgetting capability. (iv) Furthermore, more experiments are conducted to explore the proposed method, which includes utilizing a new dataset (Split TinyImageNet), conducting additional ablation studies, and visualizations.
\section{Related work}
\label{sec:relatedwork}
\subsection{Continual Learning}

There are many continual learning setups that have been discussed recently~\cite{prabhu2020gdumb}. First, continual learning can be divided into task-based and task-free setups. Task-based continual learning~\cite{delange2021continual,zhao2023adaptcl} assumes that task-ID can be available while the network in task-free continual learning~\cite{zhou2021learning} has no access to task-ID at inference time. Next, continual learning can be categorized into offline and online setups. Under the offline setup~\cite{sun2020continual}, all training samples in the current learning stage can be learned with multiple training steps. In contrast, each training sample in the streamed data is seen only once for the online setup~\cite{mai2022online}. Finally, following~\cite{koh2021online}, continual learning also can be separated into disjoint-task and blurry-task setups. The disjoint-task setup requires that the intersection of the label sets of any two learning stages is an empty set~\cite{aljundi2019task}. The blurry-task setup has been proposed to set some common samples for the training samples of each learning stage~\cite{bang2021rainbow}.

Five main directions drive recent advances in continual learning. 1) {Architecture-based methods} divide each task into a set of specific parameters of the model. They dynamically extend the model as the task increases~\cite{gao2022efficient} or gradually freeze part of parameters to overcome the forgetting problem~\cite{miao2021continual}. 2) {Regularization-based methods} store the historical information learned from old data as the prior knowledge of the network. It consolidates past knowledge by extending the loss function with an additional regularization term~\cite{kirkpatrick2017overcoming,dhar2019learning}. 3) {Replay-based methods}, which set a fixed-size memory buffer~\cite{buzzega2021rethinking} or generative model~\cite{su2019generative,dedeoglu2023continual,park2020convolutional} to store, produce, and replay historical samples during training. A large number of approaches~\cite{cha2021co2l,chaudhry2021using} have been proposed to improve the performance of replaying. Instead of replaying samples, some approaches~\cite{lao2021two,ho2023prototype} focus on the feature replaying. 4) The prompt-based methods, such as S-prompt~\cite{wang2022s}, L2P~\cite{wang2022learning}, Dualprompt~\cite{wang2022dualprompt}, and CODA-Prompt~\cite{smith2023coda} utilize prompt learning based on a pre-trained model, demonstrating better performance. 5) The optimization-based methods explicitly constrain the updating process of parameters. A typical idea is the gradient projection~\cite{wang2021training,kao2021natural,deng2021flattening} that projects the original gradient in a specified direction for updating parameters more effectively. It is different from the replay-based methods that constrain the gradient by replaying old samples.

In this work, the setup is online continual learning based on the disjoint-task setup with a task-free process. HPCR is proposed as a novel replay-based method to address the forgetting issue. Different from general replay-based methods, HPCR is used to quickly learn novel knowledge and efficiently retain historical knowledge in an online fashion.


\subsection{Online Continual Learning}
OCL is a subfield of continual learning that focuses on efficiently learning from a single pass over the online data stream. It is essential in scenarios where a model needs to continuously evolve its knowledge and adapt to new information~\cite{mai2022online}. Researchers in this field focus on developing algorithms and techniques to mitigate catastrophic forgetting, control model adaptation, and efficiently use limited resources to achieve this continuous learning. At present, the replay-based methods are the main solutions of OCL without AOP~\cite{guo2022adaptive}, which is based on orthogonal projection. Although architecture-based methods can be effective for continual learning, they are often impractical for OCL. Since it is unfeasible for large numbers of tasks~\cite{zhu2021class} by the "incremental network architectures". Meanwhile, it is known that regularization-based methods show poor performance in OCL as it is hard to design a reasonable metric to measure the importance of parameters~\cite{zhu2021class}. 

Experience replay (ER)~\cite{rolnick2019experience} that employs reservoir sampling for memory management is usually a strong strategy. Many subsequent methods have been improved based on this method and can generally be divided into three groups. 1) Some approaches belong to \textbf{memory retrieval strategy}, which usually studies how to select more important samples from the buffer for replaying. These representative methods include MIR~\cite{aljundi2019online} using increases in loss, and ASER~\cite{shim2021online} using adversarial shapley value. 2) In the meantime, some approaches~\cite{aljundi2019gradient,jin2021gradient} focus on saving more effective samples to the memory buffer, belonging to the \textbf{memory update strategy}. The core of this type of method is to use the limited samples in the buffer to approximate all previous samples as much as possible. 3) In addition to selecting important samples, it is equally important to conduct efficient anti-forgetting training. The \textbf{model update strategy} is a family of methods ~\cite{mai2021supervised,yin2021mitigating,caccia2021new,guo2022online,gu2022not,chrysakisonline,zhang2022simple,lin2023uer,wei2023online,michel2024learning,kim2024online} to improve the learning efficiency, making the model learn quickly and memory efficiently. All of them belong to the proxy-based replay manners except SCR~\cite{mai2021supervised}, which uses a contrastive-based replay manner. 

The proposed HPCR in this work is a novel model update strategy for OCL. Different from existing approaches, it is a more holistic and effective method proposed based on PCR. After conducting an in-depth analysis of PCR, HPCR improves the model’s feature extraction, generalization, and anti-forgetting capabilities with three components, making PCR better suited for OCL than previous studies.

\subsection{Deep Metric Learning}
Similar to~\cite{yao2022pcl}, our method is inspired by deep metric learning. Deep metric learning aims to measure the similarity among samples by a deep neural network while using an optimal distance metric for learning tasks~\cite{yang2018person}. Generally, the loss functions for this learning task can be categorized into pair-based and proxy-based losses. For one thing, the pair-based methods~\cite{milbich2020sharing,elezi2022group} can mine rich semantic information from anchor-to-sample relations, but converge slowly due to its high training complexity. Since contrastive-based loss is good at learning anchor-to-sample pairs, it can be regarded as a type of pair-based method. For another thing, the proxy-based methods~\cite{zheng2023deep,roth2022non} converge fast and stably, meanwhile may miss partial semantic information by learning anchor-to-proxy relations. The cross-entropy loss can be regarded as a type of proxy-based method.

The proposed HPCR in this work integrates these two types of methods. Different from existing studies, HPCR is proposed to quickly learn novel knowledge and effectively keep historical knowledge for OCL.

\begin{table*}[t]
\centering
\caption{The characteristics of the proxy-based and contrastive-based replay manners.}
\vspace{-2ex}
\label{table:characteristics}
\begin{tabular}{l|l|l|l|l}
\hline
Replay Manner  & Classifier     & Pairs of Loss      & Advantages  & Disadvantages \\ \hline
Proxy-based     & Softmax          & All Anchor-to-proxy Pairs          & \makecell[l]{Fast Convergence\\Stable}  & \makecell[l]{Bias\\Incomplete Semantic Knowledge}  \\\hline
Contrastive-based     & NCM          & Selective Anchor-to-sample Pairs          & \makecell[l]{Avoid Bias\\Complete Semantic Knowledge}  & \makecell[l]{Slow Convergence\\Unstable}  \\\hline
\end{tabular}
\vspace{-4ex}
\end{table*}

\section{Background}
\label{sec:preliminary}
\subsection{Online Continual Learning} 
OCL divides a data stream into a sequence of learning tasks as $\mathcal{D}=\{\mathcal{D}_t\}_{t=1}^T$, where $\mathcal{D}_t=\{\mathcal{X}_t\times \mathcal{Y}_t,\mathcal{C}_t\}$ contains the samples $\mathcal{X}_t$, corresponding labels $\mathcal{Y}_t$, and task-specific classes $\mathcal{C}_t$. Different tasks have no overlap in the classes. All of learned classes are denoted as $\mathcal{C}_{1:t}=\bigcup_{k=1}^t \mathcal{C}_k$. The neural network is made up of a feature extractor $\bm{z}=h(\bm{x};\bm{\Phi})$ and a softmax (proxy-based) classifier $\bm{o}=f(\bm{z};\bm{W})=[o_c]_{c=1}^{\mathcal{C}_{1:t}}$~\cite{hou2019learning}, where $o_c=\langle\bm{z},\bm{w}_c\rangle/\tau$ is the $i$th dimension value of $\bm{o}$, $\langle\cdot,\cdot\rangle$ is the cosine similarity, $\tau$ is a scale factor and $\bm{W}=[\bm{w}_1,\bm{w}_2,...,\bm{w}_c]$ contains trainable proxies of all classes. The categorical probability that sample $\bm{x}$ belongs to class $y$ is
\begin{equation}
    \label{eq:probability}
	p_y=\frac{exp(o_y)}{\sum_{c\in\mathcal{C}_{1:t}}exp(o_c)}.
\end{equation}
Generally, the model learns each mini-batch current samples ${\mathcal{B}\subset\mathcal{D}_t}$ by the following loss function
\begin{equation}
    \label{eq:oclloss}
	L=E_{(\bm{x},y)\sim{\mathcal{B}}}[-log(p_y)].
\end{equation}
In this case, the model can learn new classes well and forget old classes, resulting in the forgetting problem. To mitigate the forgetting, existing replay-based methods replay part of previous samples in two ways.

First, the proxy-based replay manners (such as ER~\cite{rolnick2019experience}) select a mini-batch of previous samples $\mathcal{B}_\mathcal{M}\subset\mathcal{M}$ to calculate proxy-based loss by
\begin{equation}
\label{eq:erloss}
\begin{split}
L_{ER}=E_{(\bm{x},y)\sim{\mathcal{B}\cup\mathcal{B}_\mathcal{M}}}[-log(\frac{exp(o_y)}{\sum_{c\in{\mathcal{C}_{1:t}}}exp(o_c)})],
\end{split}
\end{equation}
where $\mathcal{M}$ is a memory buffer to store part of previous samples. And it predicts unknown instances by a softmax classifier  
\begin{equation}
\label{eq:test}
 	\begin{split}
	 	y^* = \mathop{\arg\max}_{c}p_c,c\in C_{1:t}.
 	\end{split}
\end{equation}
Moreover, other proxy-based replay manners~\cite{ahn2021ss,caccia2021new} improve ER by selecting part of anchor-to-proxy pairs to calculate the objective function. Although effective, they can easily hurt the model's generalization ability to learn new classes.

Second, the contrastive-based replay manners (such as SCR~\cite{mai2021supervised}) replay $\mathcal{B}_\mathcal{M}\subset\mathcal{M}$ using a contrastive-based loss
\begin{equation}\small
\begin{split}
    \label{eq:scrloss}   
L_{PCR}=E_{(\bm{x},y)\sim{\mathcal{B}}\cup\mathcal{B}_\mathcal{M}}
    [\frac{-1}{|\mathcal{P}|}\sum_{p\in \mathcal{P}}
    log\frac{exp(\langle \bm{z},\bm{z}_p\rangle/\tau)}{\sum_{j\in {\mathcal{J}}}exp(\langle \bm{z},\bm{z}_j\rangle/\tau)}].
\end{split}
\end{equation} 
Different from the proxy-based loss, the selected pairs rely on the number of samples in a training batch. And it inferences by an NCM classifier
\begin{equation}
 	\begin{split}
	 	y^* = \mathop{\arg\min}_{c}\Vert\bm{z}-\bm{\mu}_c\Vert,c\in C_{1:t},
 	\end{split}
\end{equation}
To obtain the classification centers $\bm{\mu}_c$ of all classes, it has to compute the embeddings of all samples in the memory buffer before each inferring. Due to the high complexity of training, its effect is unstable and the convergence speed is slow.

\subsection{Proxy-based Contrastive Replay}
\label{sec:PCR}

Our previous work~\cite{Lin_2023_CVPR} analyzes existing studies and proposes to couple two replay manners, achieving complementary advantages. The characteristics of these two replay manners are summarized in Table~\ref{table:characteristics}. For one thing, the anchor-to-proxy pairs of proxy-based replay manner can overcome the disadvantage of contrastive-based loss. For another thing, the contrastive-based loss provides a heuristic way to select anchor-to-proxy pairs for the proxy-based replay manner. Moreover, the anchor-to-sample pairs contain richer fine-grained semantic knowledge than the anchor-to-proxy pairs.

Although there are some coupling solutions, they can not address the forgetting problem at all. For example, the most intuitive way is to add different loss functions as
\begin{equation}
\begin{split}
    \label{eq:coupleloss1}
    L_{couple}^{1}=L_{ER} + L_{SCR}.
\end{split}
\end{equation} 
It is a two-stage strategy for offline learning~\cite{li2022targeted}, which is unsuitable for OCL due to timeliness requirements. Another way is to add anchor-to-sample pairs to cross-entropy loss as
\begin{equation}
\begin{split}
    \label{eq:coupleloss2}
    &
    L_{couple}^{2}=E_{(\bm{x},y)\sim{\mathcal{B}\cup\mathcal{B}_\mathcal{M}}}\\
    &
    [-log(\frac{exp(o_y)}{\sum_{c\in{\mathcal{C}_{1:t}}}exp(o_c)+\sum_{j\in{\mathcal{J}}}exp(\langle \bm{z},\bm{z}_j\rangle/\tau)})].
\end{split}
\end{equation} 
in the work~\cite{yao2022pcl}. Both of them cannot take advantage of each replay manner to address the forgetting issue of OCL. Since they ignore that the key is to select anchor-to-proxy pairs in a contrastive way and train by the selective pairs. 

Therefore, the proxy-based contrastive replay (PCR) framework is proposed by replacing the samples of anchor-to-sample pairs with proxies in the contrastive-based loss. Specifically, the model is trained by learning samples of new classes and replaying samples of old classes in the training procedure. For each step, given current samples $\mathcal{B}$, it randomly retrieves previous samples $\mathcal{B}_\mathcal{M}$ from the memory buffer. Besides, these samples are spliced together for the batch of training. Then, the model is optimized by  
\begin{equation}
    \label{eq:pcrloss}
\begin{split}
    L_{PCR}=E_{(\bm{x},y)\sim{\mathcal{B}\cup\mathcal{B}_\mathcal{M}}}[-log(\frac{exp(o_y)}{\sum_{c\in\mathcal{C}_{1:t}}k_c exp(o_c)})].
\end{split}
\end{equation}
where $k_c$ is the number of samples belonging to class $c$ in the training batch. Different from existing studies, its way of computing categorical probability is changed for each mini-batch. Finally, it updates the memory buffer by reservoir sampling strategy, which can ensure that the probability of each sample being extracted is equal. Conveniently, the memory buffer has a fixed size, no matter how large the sample amount is.

The inference procedure is different from the training procedure. Each testing sample $\bm{x}_k$ obtains its class probability distribution by Equation (\ref{eq:probability}). And PCR performs the inference prediction to $\bm{x}_k$ with highest probability as
\begin{equation}
 	\begin{split}
	 	y_k^* = \mathop{\arg\max}_{c}p_c,y\in C_{1:t}.
 	\end{split}
\end{equation}




\section{Analysis for Proxy-based Contrastive Replay}

\subsection{Gradient Analysis}
As stated in the previous work~\cite{Lin_2023_CVPR}, gradient analysis is of great help in exploring forgetting problems and existing work. Specifically, the gradient of the softmax classifier for $\bm{x}$ is
\begin{equation}
    \label{eq:gradient_h}
    \frac{\partial L}{\partial <\bm{z},\bm{w}_c>}=\left\{
    \begin{aligned}
        (p_y-1)/\tau&,&c= y \\
        (p_c)/\tau&,&c\neq y
    \end{aligned}
    \right..
\end{equation}
Training $\bm{x}$ belongs to class $y$ provides the positive gradient $-\eta(p_y-1)/\tau$ for the proxy of class $y$ as $<\bm{z},\bm{w}_y>=<\bm{z},\bm{w}_y>-\eta(p_y-1)/\tau$, and propagates the negative gradient $-\eta(p_c)/\tau$ to the other proxies as $<\bm{z},\bm{w}_c>=<\bm{z},\bm{w}_c>-\eta(p_c)/\tau$, where the learning rate $\eta$ is positive. If training as Equation (\ref{eq:oclloss}), the proxies of new classes receive more positive gradients and the others obtain more negative gradients. It causes the proxies of new classes to be close to the samples of new classes, while the proxies of old classes are far away from the samples of new classes. Hence, it is easy to classify most samples into new classes, causing the forgetting problem. If replaying as Equation (\ref{eq:erloss}), the proxies of old classes can obtain more positive gradient, and the proxies of new classes receive more negative gradient. Although the forgetting can be mitigated, its effect is still limited by the imbalanced samples.  

To explore the reliability of PCR, we analyze its gradient propagation process. Similar to Equation (\ref{eq:gradient_h}), the gradient of the classifier for a sample $\bm{x}$ in PCR is stated as Theorem~\ref{thm1}.
\newtheorem{theorem}{\bf Theorem}
\begin{theorem}\label{thm1}
	In PCR, the gradient for all proxies is 
 \begin{equation}
    \label{eq:gradient_pcr}
    \frac{\partial L_{PCR}}{\partial <\bm{z},\bm{w}_c>}=\left\{
    \begin{aligned}
        (k_yp^*_y-1)/\tau&,&c= y \\
        (k_cp^*_c)/\tau&,&c\neq y
    \end{aligned}
    \right.,
\end{equation}
where 
\begin{equation}
\begin{split}
    p^*_y=\frac{exp(o_y)}{\sum_{c\in\mathcal{C}_{1:t}}k_j exp(o_c)}.
\end{split}
\end{equation}
\end{theorem} 
When replaying, the number of classes is relatively small and the number of samples is relatively large for new classes; the number of classes is relatively large and the number of samples is relatively small for old classes. It means that in a training batch, the number of samples for new classes tends to be multiple, while the one for old classes tends to be small, or even non-existent. Based on this situation, Equation~(\ref{eq:gradient_pcr}) demonstrates two advantages of PCR for anti-forgetting. 

For one thing, the gradient only influences the classes whose associated samples exist in the same training batch. For example, when learning a current sample $\bm{x}$ belongs to class $y$, the gradient for the proxy of class $c (c\neq y)$ is $-\eta(k_cp^*_c)/\tau$ if there are $k_c$ samples belonging to class $c$ in the same batch. Otherwise, if $k_c=0$, the proxy of class $c$ does not participate in the gradient propagation. Due to the lower probability of old classes appearing in the training batch, the proxies of old classes will receive fewer negative gradients from the new classes. Thus, the model will keep more historical knowledge. 

Another thing is that the gradient can be affected by the number of samples in a training batch. Conveniently, we define
\begin{equation}
\begin{split}
    {sum}_y=\sum_{c\in\mathcal{C}_{1:t},c\neq y}k_cexp(o_c).
\end{split}
\end{equation}
With this representation, the Equation~(\ref{eq:gradient_pcr}) can be changed to 
 \begin{equation}\small
    \label{eq:gradient_pcr_other}
    \frac{\partial L_{PCR}}{\partial <\bm{z},\bm{w}_c>}=\left\{
    \begin{aligned}
        (\frac{k_yexp(o_y)}{{sum}_y+k_yexp(o_y)}-1)/\tau&,&c= y \\
        \frac{k_cexp(o_c)}{{sum}_y+k_yexp(o_y)}/\tau&,&c\neq y
    \end{aligned}
    \right..
\end{equation}
We take the training batch including $k_i (k_i\geq1)$ samples of new class $i$ and $k_j (k_j=1)$ samples of other old classes $j$ as an example. Training new class $i$ propagates the gradient as  
\begin{equation}\small
    \frac{\partial L_{PCR}}{\partial <\bm{z},\bm{w}_c>}=\left\{
    \begin{aligned}
        -\frac{{sum}_i}{{sum}_i+k_iexp(o_i)}/\tau&,&c= i \\
        \frac{exp(o_c)}{{sum}_i+k_iexp(o_i)}/\tau&,&c= j
    \end{aligned}
    \right..
\end{equation}
At this moment, the larger the $k_i$ is, the smaller the $\frac{1}{{sum}_i+k_iexp(o_i)}$ is, and the smaller the gradient value is. It means that the larger $k_i$ can reduce the positive gradient for the proxy of new class $i$ and the negative gradient for the proxy of old classes $j$. Similarly, training any old class $j$ propagates the gradient as
\begin{equation}\small
    \frac{\partial L_{PCR}}{\partial <\bm{z},\bm{w}_c>}=\left\{
    \begin{aligned}
        (\frac{exp(o_c)}{{sum}_i+k_iexp(o_i)}-1)/\tau&,&c= j \\
        (1-\frac{{sum}_i}{{sum}_i+k_iexp(o_i)})/\tau&,&c=i
    \end{aligned}
    \right..
\end{equation}
In this situation, the larger the $k_i$ is, the smaller the $\frac{1}{{sum}_i+k_iexp(o_i)}$ is, and the larger the gradient value is. With larger $k_i$, both the positive gradient for old class $j$ and the negative gradient for the new class $i$ can be increased. Therefore, PCR helps old classes acquire some advantages in the propagation of gradient, and the gradient propagation efficiency of these selected proxies will be greatly improved.

\subsection{Limitation Analysis}
Based on the gradient analysis, we can find that PCR can help the model alleviate the phenomenon of catastrophic forgetting. It selects partial proxies instead of all to learn current samples and replay previous samples. In such a learning way, the old classes whose associated samples are not selected can avoid the bias caused by unbalanced gradient competition.

However, PCR remains suboptimal due to several unaddressed limitations. Firstly, PCR exclusively considers the relationships of anchor-to-proxy but ignores the relations of anchor-to-sample, leading to the model losing some important semantic information. Anchor-to-sample pairs play a crucial role in enhancing the model's feature extraction capability in large training batches, a vital aspect of PCR. Secondly, the simplistic setting of the temperature coefficient $\tau$ in PCR lacks sophistication. As the temperature coefficient influences PCR's gradient propagation process, employing a static constant is insufficient to ensure the model's generalization ability. Thirdly, those old classes that have been selected to learn together with new classes still face the issue of bias in the training process. It means that the model's anti-forgetting ability still can be further improved. In a word, PCR can be improved into a more holistic and effective approach.

\section{Holistic Proxy-based Contrastive Replay}
\label{sec:PCR+}
In this work, we develop a more comprehensive and effective method called holistic proxy-based contrastive replay (HPCR). HPCR consists of a contrastive component, a temperature component, and a distillation component, resolving the limitations of PCR in sequence. 

\subsection{Contrastive Component}
 The contrastive component conditionally incorporates the anchor-to-sample pairs to PCR and improves the model's ability of feature extraction. The anchor-to-sample pairs contain richer fine-grained semantic information compared to anchor-to-proxy pairs. Although extracting the relations of anchor-to-sample is easily limited by the number of samples in a training batch, it can still play a crucial role when the batch size is sufficient. Hence, the contrastive component of HPCR is denoted as 

\begin{equation}\small
    \label{eq:pcr+}
\begin{split}
    &
    L_{PCR_{C}}=E_{(\bm{x},y)\sim{\mathcal{B}\cup\mathcal{B}_\mathcal{M}}}[\frac{-1}{|\mathcal{P}|}\sum_{p\in\mathcal{P}}
    \\
    &
    log(\frac{exp(o_y)+s(N)\cdot exp(\langle \bm{z},\bm{z}_p\rangle/\tau)}{\sum_{c\in\mathcal{C}_{1:t}}k_c exp(o_c)+\sum_{j\in\mathcal{J}} s(N)\cdot exp(\langle\bm{z},\bm{z}_j\rangle/\tau)})].
\end{split}
\end{equation}
Here, $s(N)$ is a self-defined step function used to control the importance of anchor-to-sample pairs, and it is defined as
\begin{equation}
\label{eq:step_function}
    s(N)=\left\{
    \begin{aligned}
        0,\quad&N < N_{min}\\
        1,\quad&N \geq N_{min}
    \end{aligned}
    \right..
\end{equation}
$N$ is the number of samples in the current training batch, and $N_{min}$ is a hyper-parameter used to determine whether to use anchor-to-sample pairs. When the size of the training batch is small, the correlation between samples is prone to noise affecting the performance of the model. Therefore, their importance should be set to 0. At this point, $\rm PCR_C$ degenerates into PCR; otherwise, the importance of these samples should be set to 1 to improve the model's feature extraction ability. Compared to the PCR, $\rm PCR_C$ can not only mine the relations of anchor-to-proxy, but also explore the relations of anchor-to-sample. It increases inter-class distance as well as decreases intra-class distance.
\begin{figure*}[t]
\centering
    \begin{minipage}[t]{0.44\linewidth}
        \centering
        \includegraphics[scale=0.22]{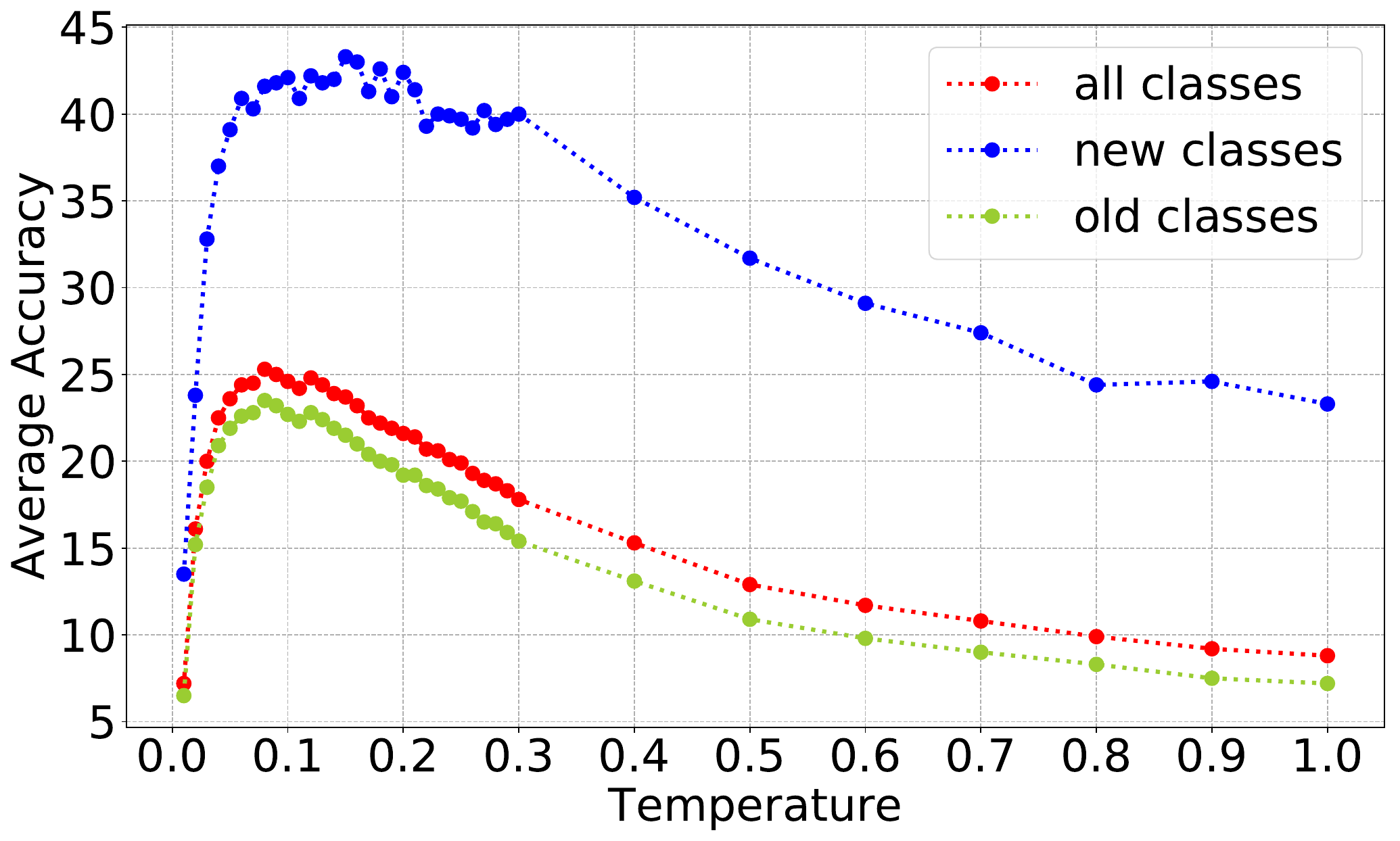}
        {(a) The performance on different classes}
    \end{minipage}
    \begin{minipage}[t]{0.44\linewidth}
        \centering
        \includegraphics[scale=0.22]{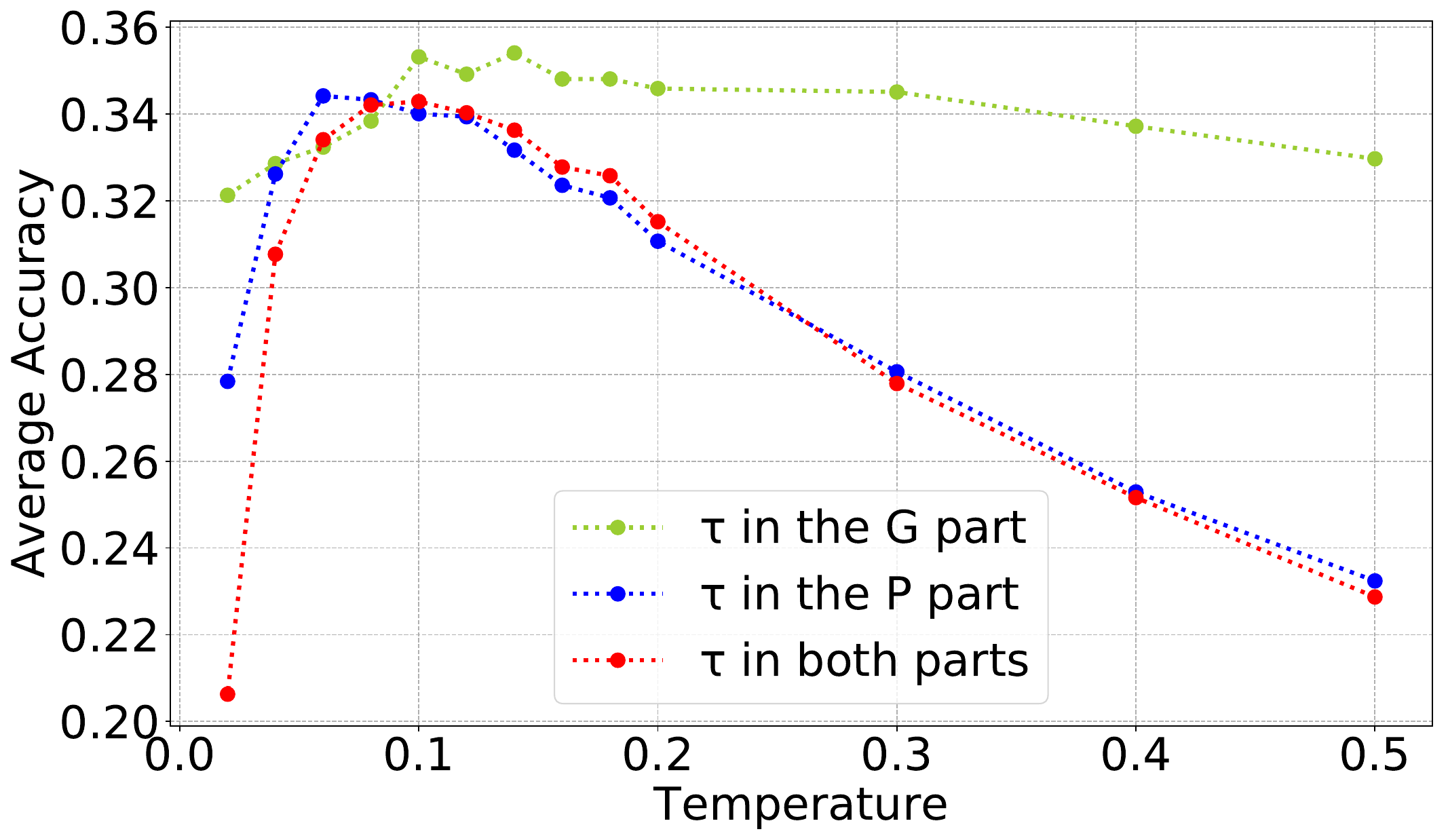}
        {(b) The performance on different parts}
    \end{minipage}
\centering
\vspace{-1ex}
\caption{The performance of PCR on Split CIFAR100 (buffer size=1000) with different value of $\tau$. }
\label{fig:tau}
\vspace{-2ex}
\end{figure*}

\subsection{Temperature Component}

\subsubsection{The Analysis of Temperature Coefficient}
In Equation (\ref{eq:gradient_pcr}), the gradient propagation is influenced by $\tau$, which is called the temperature coefficient. In a proxy-based classifier, the norm of vectors has been proved to be unfriendly to continual learning and thus abandoned~\cite{hou2019learning}. However, the retained similarity of vectors $\langle\bm{z},\bm{w}\rangle$ has a value range of $[-1,1]$, which brings certain difficulties to the optimization of the model. Hence, $\tau$ in $p^*_c$ is used to control the strength of $\langle\bm{z},\bm{w}\rangle$, which is vital to find the optimal solution. 

Inspired by~\cite{wang2021understanding}, we define a relative penalty $r(\langle\bm{z},\bm{w}_c\rangle)$ on negative proxy $\bm{w}_c(c\neq y)$ for the sample $(\bm{x},y)$ by
\begin{equation}
    \begin{split}
    r(\langle\bm{z},\bm{w}_c\rangle)=\frac{|\frac{\partial L}{\partial <\bm{z},\bm{w}_c>}|}{|\frac{\partial L}{\partial <\bm{z},\bm{w}_y>}|}=\frac{k_cexp(o_c)}{\sum_{j\neq y}k_jexp(o_j)}, c\neq y.
    \end{split}
\end{equation}
It can reflect the strength of penalties on all negative samples. As $\tau$ increases, the distribution of $r(\langle\bm{z},\bm{w}_c\rangle)$ is relatively uniform, and the model tends to treat all negative proxies with the same strength. On the contrary, the model focuses on the negative proxies with higher similarities when $\tau$ decreases.

Furthermore, within a reasonable interval, smaller $\tau$ is more suitable for replaying old classes, while larger $\tau$ is more effective for learning new classes in OCL. Generally, the number of samples in old classes is small, while the number of samples in new classes is large in OCL. It leads to the samples of old classes being more likely to become negative samples with shorter distances, while the samples of new classes tend to become negative samples with larger distances. As a result, large $\tau$ will prompt the model to pay attention to the samples of new classes with larger distances, making it easy to ignore samples from old classes. In contrast, small $\tau$ will urge the model to focus on the samples of old classes with shorter distances, resulting in the information loss from new classes.

Finally, as seen in Equation (\ref{eq:gradient_pcr}), $\tau$ affects the process in two parts. First, it directly changes the value of the gradient by ${1}/{\tau}$ (denoted as G part). Second, it smooths the probability distribution $p^*_c$ of samples to implicitly adjust the gradient (denoted as P part). Moreover, the P part has a more complex impact than the G part due to its stronger non-linearity. 

\subsubsection{The Influence of Temperature Coefficient}
The performance of PCR would be highly sensitive to $\tau$. To validate this viewpoint, we conduct experiments for PCR with different values of $\tau$, and the results are shown in Figure~\ref{fig:tau}.
In addition to the final accuracy rate of all classes, we also report the ones of new and old classes separately. In Figure~\ref{fig:tau}(a), we can make the following observations: 1) The performances of all, old, and new classes are sensitive to $\tau$. For example, when $\tau$ changes from 0.01 to 0.2, the performance of PCR in all aspects would first increase and then decrease. 2) The sensitivity of old and new classes w.r.t. $\tau$ is different, and their best $\tau$ are different. The performance on new classes reaches the best result when $\tau\in[0.1,0.2]$ while the performance on old classes reaches the best result when $\tau\in[0.0,0.1]$. 

At the same time, we also demonstrate the performance when different $\tau$ are taken in different parts. In Figure~\ref{fig:tau}(b), the lines of green, blue, and red denote the performance when $\tau$ changes in the G part, the P part, and both parts. For the G part, we keep $\tau=0.09$ for $p^*_c$ and only change the value of $\tau$ for ${1}/{\tau}$. On the contrary, $\tau$ is set as 0.09 for ${1}/{\tau}$ and changed from 0.0 to 0.5 in the P part. Comparing the results of the two cases, we find that the performance of the model is more sensitive to changes in $\tau$ in the G part. Meanwhile, the model exhibits greater fluctuations when $\tau$ reaches the optimal value in the P part. Although the performance of the P part appears stable with changes in $\tau$, the accuracy of each class actually changes. Conveniently, we report the best $\tau$ for each class on Split CIFAR10 in the P part when the buffer size is 100. As stated in Table~\ref{table:tau}, to achieve the highest accuracy, the value of $\tau$ for each class is different. In a word, the setting of $\tau$ in the P part is sensitive, and a static value is more suitable for it. While a dynamic value is more suitable for the G part since it allows the model to consider different classes. Therefore, the setting of $\tau$ for these two parts should be decoupled.

\begin{table}[t]
\renewcommand\tabcolsep{4pt}
\centering
\caption{The best $\tau$ for each class on Split CIFAR10 (buffer size=100).}
\vspace{-1ex}
\label{table:tau}
\begin{tabular}{l|llllllllll}
\hline
class & 0    & 1    & 2    & 3    & 4    & 5    & 6    & 7    & 8    & 9    \\ \hline
type  & old  & old  & old  & old  & old  & old  & old  & old  & new  & new  \\
$\tau$     & 0.01 & 0.14 & 0.16 & 0.17 & 0.12 & 0.20 & 0.06 & 0.09 & 0.14& 0.2 \\ \hline
\end{tabular}
\vspace{-4ex}
\end{table}

\subsubsection{Decoupled Temperature Coefficient}
To this end, we set different values of $\tau$ for the P and G parts separately. For one thing, since the performance of the model is more sensitive to the P part, we set it to a static constant value. Meanwhile, we obtain a moderate value to balance the model's attention to new and old classes using the grid search method. For another thing, due to the different optimal $\tau$ for each class, we set a dynamic one for the G part. However, it is highly difficult to find an optimal $\tau$ for each class at any time, since the real-time requirement of OCL is high. In such a situation, the dynamic temperature is set as a robust function independent of the class, and the function is denoted as
\begin{equation}
\label{eq:tau_t}
 	\begin{split}
	 	\tau(s) = (\tau_{max}-\tau_{min})\times(1+cos(2\pi s/S))/2+\tau_{min},
 	\end{split}
\end{equation}
where $s$ is the current step, $S$ is the cycle length, $\tau_{max}$ is the upper bound, and $\tau_{min}$ is the lower bound. With such a periodic dynamic function, the model can select a wide range of different temperature values during the OCL process. It improves the accuracy of each class and thus enhances the overall generalization ability of the model. Therefore, the Equation (\ref{eq:pcr+}) can be improved to
\begin{equation}\small
\label{eq:pcr++}
\begin{split}
&
L_{PCR_{CT}}=E_{(\bm{x},y)\sim{\mathcal{B}\cup\mathcal{B}_\mathcal{M}}}[\frac{-1}{|\mathcal{P}|}\sum_{p\in\mathcal{P}}\frac{\tau}{\tau(s)}
    \\
    &
    log(\frac{exp(o_y)+s(N)\cdot exp(\langle \bm{z},\bm{z}_p\rangle/\tau)}{\sum_{c\in\mathcal{C}_{1:t}}k_c exp(o_c)+\sum_{j\in\mathcal{J}} s(N)\cdot exp(\langle\bm{z},\bm{z}_j\rangle/\tau)})].
\end{split}
\end{equation}
Moreover, its corresponding gradient is transformed into

\begin{equation}
    \label{eq:gradient_pcr++}
    \frac{\partial L_{PCR_{CT}}}{\partial <\bm{z},\bm{w}_c>}=\left\{
    \begin{aligned}
        (k_yp^*_y-1)/\tau(s)&,&c= y \\
        (k_cp^*_c)/\tau(s)&,&c\neq y
    \end{aligned}
    \right..
\end{equation}

\subsection{Distillation Component}
As a method of knowledge distillation~\cite{hinton2015distilling}, DER~\cite{buzzega2020dark} tries to retain the logits output obtained from the model during learning as historical knowledge for sample replaying. This is to improve the diversity of historical knowledge and avoid the problem of knowledge singularity caused by saving old models to store historical knowledge. However, existing DER has to use all proxies to calculate a loss function of distillation. If distilling as DER, the proxies excluded by PCR will continue to participate in gradient propagation. 

To avoid this situation, we propose a novel distillation method called proxy-based contrastive distillation (PCD). Similar to PCR, the categorical probability in PCD only uses the proxies that appear in the training batch. Specifically, when saving a current sample into the memory buffer, we not only save its instance $\bm{x}$ and label $y$, but also store its logits output of classifier $\bm{o}^*$. If the sample is selected as a previous sample to replay, its $\bm{o}^*$ can be used to keep richer historical knowledge. Hence, we can calculate the PCD loss by minimizing the Euclidean distance between the weighted logits as

\begin{equation}
\label{eq:pcd}
\begin{split}
    L_{PCD}=E_{(\bm{x},y,\bm{o}^*)\sim{\mathcal{B}_\mathcal{M}}}[-\sum_{c}^{\mathcal{C}_{1:t}}k_c(o_c-o^*_c)^2].
\end{split}
\end{equation}

In addition to the distribution distillation of previous samples, relation distillation between samples is also necessary. Since the relation of samples not only reflects intra-class information, but also includes inter-class correlation. Hence, we propose to compute sample-based contrastive distillation (SCD) by $\bm{z}^*$ in a training batch as
\begin{equation}\small
\label{eq:scd}
\begin{split}
    &
    L_{SCD}=E_{(\bm{x},y,\bm{z}^*)\sim{\mathcal{B}_\mathcal{M}}}\\
    &
    [-\sum_{i\in\mathcal{J}}(\frac{exp(\langle\bm{z},\bm{z}_i\rangle/\tau)}{\sum_{j\in\mathcal{J}}exp(\langle\bm{z},\bm{z}_j\rangle/\tau)}log\frac{exp(\langle\bm{z}^*,\bm{z}^*_i\rangle/\tau)}{\sum_{j\in\mathcal{J}}exp(\langle\bm{z}^*,\bm{z}^*_j\rangle/\tau})],
\end{split}
\end{equation}
where $\mathcal{J}$ is the indices set of samples except for anchor $\bm{x}$ in the same batch $\mathcal{B}_\mathcal{M}$. With these two distillation loss functions, the model can transfer more positive gradients to the proxies of old classes and more negative gradients to the proxies of new classes. As a result, the model's ability of anti-forgetting can be improved.

\begin{algorithm}[t]
\caption{Holistic Proxy-based Contrastive Replay}
\label{alg:hpcr}
\textbf{Input}: Dataset $D$, Learning Rate $\lambda$, Scale factor $\tau$\\
\textbf{Output}:Network Parameters $\theta$\\
\textbf{Initialize}:Memory Buffer $\mathcal{M}\leftarrow\{\}$, Network Parameters $\bm{\theta}=\{\bm{\Phi},\bm{W}\}$
\begin{algorithmic}[1]
\FOR{$t\in\{1,2,...,T\}$}
\STATE //$Training\ Procedure$
    \FOR{mini-batch $\mathcal{B}\subset D_t$}
        \FOR{$(\bm{x},y) \in \mathcal{B}$}
            \STATE $\bm{z}=h(\bm{x};\bm{\Phi})$ 
            \STATE $\bm{o}=f(\bm{z};\bm{W})$
            \STATE $(\bm{x},y)\leftarrow(\bm{x},y,\bm{o},\bm{z})$
        \ENDFOR
        \STATE$\mathcal{B}_\mathcal{M}\leftarrow RandomRetrieval(\mathcal{M})$
        \STATE$\mathcal{B}\cup\mathcal{B}_\mathcal{M}\leftarrow Concat([\mathcal{B},\mathcal{B}_\mathcal{M}])$
        \STATE$L_{PCR_{CT}}\leftarrow$Equation~(\ref{eq:pcr++})
        \STATE$L_{PCD}\leftarrow$Equation~(\ref{eq:pcd})
        \STATE$L_{SCD}\leftarrow$Equation~(\ref{eq:scd})
        \STATE$L\leftarrow$Equation~(\ref{eq:hpcr})
        \STATE$\theta\leftarrow\theta+\lambda\nabla_{\theta}L$
        \STATE$\mathcal{M}\leftarrow ReservoirUpdate(\mathcal{M},\mathcal{B})$
    \ENDFOR
\STATE //$Inference\ Procedure$
\STATE $m\leftarrow number\ of\ testing\ samples$

    \FOR{$k\in\{1,2,...,m\}$}
    \STATE $y_k^*\leftarrow \mathop{\arg\max}_{c}p_c,c\in C_{1:t}$
    \ENDFOR
\STATE \textbf{return} $\theta$
\ENDFOR
\end{algorithmic}
\end{algorithm}

\subsection{Overall Procedure}
Combining these three novel components, PCR will be enhanced as HPCR. Compared to PCR, HPCR will greatly improve the model's capability of feature extraction, generalization, and anti-forgetting in the meantime. Its objective function is denoted as

\begin{equation}
\label{eq:hpcr}
\begin{split}
    &
    L_{HPCR}=L_{PCR_{CT}}+\alpha L_{PCD}+\beta L_{SCD},
\end{split}
\end{equation}
where $\alpha$ and $\beta$ are hyper-parameters of scale for PCD and SCD, respectively. Furthermore, the whole training and inference procedures of HPCR are summarized in Algorithm \ref{alg:hpcr}. The main differences between PCR and HPCR lie in the training procedure. The first difference is in the memory buffer. In addition to saving the old samples and their corresponding labels, HPCR also saves the logits and feature vectors obtained during the training process of the old samples. The second one is in the loss function. The objective function of HPCR is more integrated, which is conducive to comprehensively improving the overall performance of the model.

\begin{table*}[t]\small
\renewcommand\tabcolsep{2pt}
\centering
\caption{Final Accuracy Rate (↑). The best scores for our methods are in boldface, and the best scores for baselines are underlined.}
\vspace{-2ex}
\label{tableaccuracy}
\begin{tabular}{l|ccc|ccc|ccc|ccc}
\hline
Datasets & \multicolumn{3}{c|}{Split CIFAR10 (\%)} & \multicolumn{3}{c|}{Split CIFAR100 (\%)}& \multicolumn{3}{c|}{Split MiniImageNet (\%)} & \multicolumn{3}{c}{Split TinyImageNet (\%)}\\ \hline
Buffer & \multicolumn{1}{c|}{100} & \multicolumn{1}{c|}{200} & \multicolumn{1}{c|}{500} & \multicolumn{1}{c|}{1000}    & \multicolumn{1}{c|}{2000} & \multicolumn{1}{c|}{5000} & \multicolumn{1}{c|}{1000} & \multicolumn{1}{c|}{2000}     &\multicolumn{1}{c|}{5000} & \multicolumn{1}{c|}{2000} & \multicolumn{1}{c|}{4000} & 10000     \\ \hline
IID          & \multicolumn{3}{c|}{58.1\scriptsize±2.5}                                  & \multicolumn{3}{c|}{17.3\scriptsize±0.8}                                     & \multicolumn{3}{c|}{18.2\scriptsize±1.1} & \multicolumn{3}{c}{17.3\scriptsize±1.7}                                    \\
IID++~\cite{caccia2021new}          & \multicolumn{3}{c|}{64.2\scriptsize±2.1}                                  & \multicolumn{3}{c|}{23.5\scriptsize±0.8}                                     & \multicolumn{3}{c|}{20.7\scriptsize±1.0} & \multicolumn{3}{c}{19.1\scriptsize±1.3}                                    \\
FINE-TUNE          & \multicolumn{3}{c|}{17.9\scriptsize±0.4}                                  & \multicolumn{3}{c|}{5.9\scriptsize±0.2}                                     & \multicolumn{3}{c|}{4.3\scriptsize±0.2} & \multicolumn{3}{c}{4.3\scriptsize±0.2}                                    \\\hline

ER~\cite{rolnick2019experience} (NeurIPS2019) & 33.8\scriptsize±3.2& 41.7\scriptsize±2.8                 & 46.0\scriptsize±3.5                 & 17.6\scriptsize±0.9                  & 19.7\scriptsize±1.6                  & 20.9\scriptsize±1.2 & 13.4\scriptsize±0.9                   & 16.5\scriptsize±0.9                  & 16.2\scriptsize±1.7 & 6.1\scriptsize±0.5                   & 8.5\scriptsize±0.7                  & 8.9\scriptsize±0.6\\
GSS~\cite{aljundi2019gradient} (NeurIPS2019) & 23.1\scriptsize±3.9& 28.3\scriptsize±4.6                 & 36.3\scriptsize±4.1                 & 16.9\scriptsize±1.4                  & 19.0\scriptsize±1.8                  & 20.1\scriptsize±1.1 & 13.9\scriptsize±1.0                   & 14.6\scriptsize±1.1                  & 15.5\scriptsize±0.9 & /                   & /                  & /\\
GMED~\cite{jin2021gradient} (NeurIPS2021) & 32.8\scriptsize±4.7& 43.6\scriptsize±5.1                 & 52.5\scriptsize±3.9                 & 18.8\scriptsize±0.7                  & 21.1\scriptsize±1.2                  & 23.0\scriptsize±1.5 & 15.3\scriptsize±1.3                   & 18.0\scriptsize±0.8                  & 19.6\scriptsize±1.0 & 7.0\scriptsize±0.9                   & 10.2\scriptsize±0.7                  & 11.3\scriptsize±1.2\\
MIR~\cite{aljundi2019online} (NeurIPS2019)& 34.8\scriptsize±3.3& 40.3\scriptsize±3.3                 & 42.6\scriptsize±1.7                 & 18.1\scriptsize±0.7                  & 20.3\scriptsize±1.6                  & 21.6\scriptsize±1.7 & 14.8\scriptsize±1.1                   & 17.2\scriptsize±0.8                  & 17.2\scriptsize±1.2 & 4.9\scriptsize±0.6                   & 6.3\scriptsize±0.6                  & 6.4\scriptsize±0.7\\
ASER~\cite{shim2021online} (AAAI2021) & 33.7\scriptsize±3.7& 31.6\scriptsize±3.4                 & 42.1\scriptsize±3.0                 & 16.1\scriptsize±1.1                  & 17.7\scriptsize±0.7                  & 18.9\scriptsize±1.0 & 13.8\scriptsize±0.9                   & 16.1\scriptsize±0.9                  & 18.1\scriptsize±1.1 & 5.3\scriptsize±0.3                   & 9.6\scriptsize±0.8                  & 8.1\scriptsize±0.8\\
A-GEM~\cite{chaudhry2018efficient} (ICLR2019)& 17.5\scriptsize±1.7& 17.4\scriptsize±2.1                 & 17.9\scriptsize±0.7                 & 5.6\scriptsize±0.5                  & 5.4\scriptsize±0.7                  & 4.6\scriptsize±1.0 & 4.7\scriptsize±1.1                   & 5.0\scriptsize±2.3                  & 4.8\scriptsize±0.8 & /                   & /                  & /\\
ER-WA~\cite{zhao2020maintaining} (CVPR2020)& 36.9\scriptsize±2.9& 42.5\scriptsize±3.4                 & 48.6\scriptsize±2.7                 & 21.7\scriptsize±1.2                  & 23.6\scriptsize±0.9                  & 24.0\scriptsize±1.8 & 17.1\scriptsize±0.9                   & 18.9\scriptsize±1.4                  & 18.5\scriptsize±1.5 & 9.2\scriptsize±0.7                   & 11.6\scriptsize±1.3                  & 11.4\scriptsize±1.6\\
DER++~\cite{buzzega2020dark} (NeurIPS2020)& 40.9\scriptsize±1.4& 45.3\scriptsize±1.7                 & 52.8\scriptsize±2.2                 & 17.2\scriptsize±1.1                  & 19.5\scriptsize±1.2                  & 20.2\scriptsize±1.3 & 14.8\scriptsize±0.7                   & 16.1\scriptsize±1.3                  & 15.5\scriptsize±1.3 & 6.8\scriptsize±0.4                   & 8.7\scriptsize±0.7                  & 9.2\scriptsize±0.7\\
SS-IL~\cite{ahn2021ss} (ICCV2021) & 36.8\scriptsize±2.1 & 42.2\scriptsize±1.4                 &  44.8\scriptsize±1.6                 & 21.9\scriptsize±1.1                  & 24.5\scriptsize±1.4                  & 24.7\scriptsize±1.0 & 19.7\scriptsize±0.9                   & 21.7\scriptsize±1.0                  & 24.4\scriptsize±1.6 &\underline{13.2\scriptsize±0.8}                   & \underline{15.2\scriptsize±1.0}                  & \underline{18.7\scriptsize±0.7}\\
SCR~\cite{mai2021supervised} (CVPR-W2021) & 35.0\scriptsize±2.9& 45.4\scriptsize±1.0                 & 55.7\scriptsize±1.6                 & 16.2\scriptsize±1.3                  & 18.2\scriptsize±0.8                  & 19.3\scriptsize±1.0 & 14.7\scriptsize±1.9                   & 16.8\scriptsize±0.6                  & 18.6\scriptsize±0.5 & 9.9\scriptsize±0.4                   & 12.6\scriptsize±0.6                  & 11.1\scriptsize±0.5\\
ER-DVC~\cite{gu2022not} (CVPR2022) & 36.3\scriptsize±2.6 & 45.4\scriptsize±1.4                 &  50.6\scriptsize±2.9                & 19.7\scriptsize±0.7                  & 22.1\scriptsize±0.9                  & 24.1\scriptsize±0.8 & 15.4\scriptsize±0.7                   & 17.2\scriptsize±0.8                  & 19.1\scriptsize±0.9 & 7.6\scriptsize±0.5                   & 9.9\scriptsize±0.7                  & 10.4\scriptsize±0.7\\
OCM~\cite{guo2022online} (ICML2022) & {44.4\scriptsize±1.5} & {49.9\scriptsize±1.8}                 &  {55.8\scriptsize±2.3}                & 20.6\scriptsize±1.2                  & 22.1\scriptsize±1.0                  & 22.7\scriptsize±1.4 & 13.6\scriptsize±0.7                   & 16.5\scriptsize±0.5                  & 19.2\scriptsize±0.7 & 8.6\scriptsize±0.8                   & 11.9\scriptsize±0.9                  & 12.1\scriptsize±0.6\\
ER-ACE~\cite{caccia2021new} (ICLR2022)   & 44.3\scriptsize±1.5 & 49.7\scriptsize±2.4                & 54.9\scriptsize±1.4                 & {23.1\scriptsize±0.8}                &{24.8\scriptsize±0.9}                  & {27.0\scriptsize±1.2} & {20.3\scriptsize±1.3}                   & {24.8\scriptsize±1.1}                  & {26.2\scriptsize±1.0} & 9.5\scriptsize±0.5                   & 13.7\scriptsize±0.7                  & 18.2\scriptsize±0.5\\
OBC~\cite{chrysakisonline} (ICLR2023)   & 40.5\scriptsize±2.1  & 46.4\scriptsize±1.6  & 53.4\scriptsize±2.3                 & 22.1\scriptsize±0.6  & 24.0\scriptsize±1.3  & 26.3\scriptsize±1.0 & 16.4\scriptsize±1.4  & 19.5\scriptsize±1.5  & 21.6\scriptsize±1.4 & 9.6\scriptsize±0.5                   & 11.4\scriptsize±0.9                  & 14.6\scriptsize±1.1\\
OnPro~\cite{wei2023online} (ICCV2023)   & \underline{46.0\scriptsize±1.6}  & \underline{52.9\scriptsize±2.0}  & \underline{59.5\scriptsize±0.7}                 & 17.4\scriptsize±0.8  & 19.4\scriptsize±0.4  & 21.6\scriptsize±0.6 & 13.7\scriptsize±0.9  & 16.8\scriptsize±0.7  & 18.1\scriptsize±1.1 & 10.2\scriptsize±0.8                   & 13.6\scriptsize±0.7                  & 16.5\scriptsize±0.4\\
UER~\cite{lin2023uer} (ACMMM2023)   & {41.5\scriptsize±1.4}& {49.2\scriptsize±1.7}                & {55.8\scriptsize±1.9}                 & \underline{24.6\scriptsize±0.8}                &\underline{27.0\scriptsize±0.5}                  & \underline{29.6\scriptsize±1.1} & \underline{21.9\scriptsize±1.3}                   & \underline{25.1\scriptsize±1.1}                  & \underline{27.5\scriptsize±1.1} & 10.6\scriptsize±0.5                   & 13.8\scriptsize±0.7                  & 17.2\scriptsize±0.6\\
\hline
HPCR (Algorithm \ref{alg:hpcr})   & \textbf{48.3\scriptsize±1.5}& \textbf{53.4\scriptsize±1.4}                & \textbf{60.1\scriptsize±1.1}                 & \textbf{29.1\scriptsize±0.7}                &\textbf{30.7\scriptsize±0.5}                 & \textbf{33.7\scriptsize±0.6} & \textbf{27.1\scriptsize±0.6}                   & \textbf{29.9\scriptsize±0.7}                  & \textbf{31.3\scriptsize±0.7} & \textbf{16.4\scriptsize±0.3}                   & \textbf{19.5\scriptsize±0.8}                  & \textbf{22.1\scriptsize±0.5}\\
$\hookrightarrow$PCR    & 45.4\scriptsize±1.3& 50.3\scriptsize±1.5                & 56.0\scriptsize±1.2                 & 25.6\scriptsize±0.6                &27.4\scriptsize±0.6                 & 29.3\scriptsize±1.1 & 24.2\scriptsize±0.9                   & 27.2\scriptsize±1.2                  & 28.4\scriptsize±0.9 & 12.2\scriptsize±0.9                   & 17.4\scriptsize±0.7                  & 19.6\scriptsize±0.8\\
\hline
\end{tabular}
\vspace{-2ex}
\end{table*}

\begin{table*}[t]\small
\renewcommand\tabcolsep{1pt}
\centering
\caption{Final Accuracy Rate (↑) on Split CIFAR100 under the experimental setting of ~\cite{guo2022online}.}
\vspace{-2ex}
\label{moreaccuracy}
\begin{tabular}{l|ccc|ccc|ccc|ccc|ccc}
\hline
Methods & \multicolumn{3}{c|}{SCR~\cite{guo2022online} (\%)} & \multicolumn{3}{c|}{OCM~\cite{guo2022online} (\%)}& \multicolumn{3}{c|}{OnPro~\cite{wei2023online} (\%)} & \multicolumn{3}{c|}{PCR (\%)} & \multicolumn{3}{c}{HPCR (\%)}\\ \hline
Buffer & \multicolumn{1}{c|}{1000} & \multicolumn{1}{c|}{2000} & \multicolumn{1}{c|}{5000} & \multicolumn{1}{c|}{1000}    & \multicolumn{1}{c|}{2000} & \multicolumn{1}{c|}{5000} & \multicolumn{1}{c|}{1000} & \multicolumn{1}{c|}{2000}     &\multicolumn{1}{c|}{5000} & \multicolumn{1}{c|}{1000} & \multicolumn{1}{c|}{2000} &\multicolumn{1}{c|}{5000} & \multicolumn{1}{c|}{1000} & \multicolumn{1}{c|}{2000} & 5000     \\ \hline
$A_T$   & 26.5\scriptsize±0.2& 31.6\scriptsize±0.5                 & 36.5\scriptsize±0.2                 & 28.1\scriptsize±0.3                  & 35.0\scriptsize±0.4                  & 42.4\scriptsize±0.5 & 30.0\scriptsize±0.4                   & 35.9\scriptsize±0.6                  & 42.7\scriptsize±0.9 & 29.3\scriptsize±0.6                   & 36.3\scriptsize±0.9                  & 46.5\scriptsize±0.8 & \textbf{33.6\scriptsize±0.6}                   & \textbf{40.5\scriptsize±1.5}                  & \textbf{49.1\scriptsize±1.2} \\
\hline
\end{tabular}
\vspace{-2ex}
\end{table*}

\begin{table*}[t]\small
\renewcommand\tabcolsep{2pt}
\centering
\caption{Averaged Anytime Accuracy (↑). The best scores for our methods are in boldface, and the best for baselines are underlined.}
\vspace{-2ex}
\label{tableaaa}
\begin{tabular}{l|ccc|ccc|ccc|ccc}
\hline
Datasets & \multicolumn{3}{c|}{Split CIFAR10 (\%)} & \multicolumn{3}{c|}{Split CIFAR100 (\%)}& \multicolumn{3}{c|}{Split MiniImageNet (\%)} & \multicolumn{3}{c}{Split TinyImageNet (\%)}\\ \hline
Buffer & \multicolumn{1}{c|}{100} & \multicolumn{1}{c|}{200} & \multicolumn{1}{c|}{500} & \multicolumn{1}{c|}{1000}    & \multicolumn{1}{c|}{2000} & \multicolumn{1}{c|}{5000} & \multicolumn{1}{c|}{1000} & \multicolumn{1}{c|}{2000}     &\multicolumn{1}{c|}{5000} & \multicolumn{1}{c|}{2000} & \multicolumn{1}{c|}{4000} & 10000     \\ \hline
ER~\cite{rolnick2019experience} (NeurIPS2019) & 52.7\scriptsize±2.7& 52.1\scriptsize±3.1                 & 57.0\scriptsize±2.7                 & 25.5\scriptsize±1.1                  & 26.3\scriptsize±1.4                  & 25.3\scriptsize±1.3 & 20.1\scriptsize±0.9                   & 22.2\scriptsize±1.2                  & 19.0\scriptsize±1.5 & 11.9\scriptsize±0.9                   & 12.8\scriptsize±0.9                  & 13.0\scriptsize±0.9\\
MIR (NeurIPS2019)& 49.5\scriptsize±3.4& 50.8\scriptsize±4.2                 & 53.4\scriptsize±4.0                 & 25.8\scriptsize±0.9                  & 26.5\scriptsize±2.1                  & 26.3\scriptsize±1.8 & 20.9\scriptsize±1.6                   & 22.8\scriptsize±1.5                  & 20.5\scriptsize±1.3 & 9.1\scriptsize±0.6                   & 9.6\scriptsize±0.6                  & 9.9\scriptsize±0.7\\
ER-WA~\cite{zhao2020maintaining} (CVPR2020)& 54.1\scriptsize±4.9& 54.4\scriptsize±5.2                 & 58.1\scriptsize±4.2                 & 29.4\scriptsize±1.7                  & 29.3\scriptsize±1.9                  & 27.6\scriptsize±2.1 & 23.8\scriptsize±1.1                   & 23.5\scriptsize±1.3                  & 21.7\scriptsize±1.2 & 14.8\scriptsize±0.5                   & 14.8\scriptsize±1.5                  & 13.6\scriptsize±0.8\\
DER++~\cite{buzzega2020dark} (NeurIPS2020)& 57.1\scriptsize±2.6& 58.0\scriptsize±2.5                 & 62.1\scriptsize±4.3                 & 26.7\scriptsize±1.4                  & 26.0\scriptsize±1.5                  & 26.2\scriptsize±1.9 & 21.2\scriptsize±0.9                   & 21.3\scriptsize±1.3                  & 20.9\scriptsize±1.7 & 13.3\scriptsize±0.7                   & 13.2\scriptsize±0.9                  & 13.7\scriptsize±1.2\\
SS-IL~\cite{ahn2021ss} (ICCV2021) & 50.5\scriptsize±1.6& 52.9\scriptsize±1.2                 & 54.5\scriptsize±1.2                 & 25.1\scriptsize±1.4                  & 27.2\scriptsize±1.2                  & 26.8\scriptsize±1.1 & 22.9\scriptsize±1.5                   & 22.8\scriptsize±0.9                  & 25.5\scriptsize±1.8 & 17.1\scriptsize±0.8                   & 18.6\scriptsize±0.9                  & 19.3\scriptsize±0.9\\
SCR~\cite{mai2021supervised} (CVPR-W2021) & 56.7\scriptsize±2.8& 59.4\scriptsize±3.3                 & 67.9\scriptsize±2.9                 & 25.1\scriptsize±2.6                  & 28.7\scriptsize±2.4                  & 27.5\scriptsize±1.5 & 22.9\scriptsize±1.3                   & 24.5\scriptsize±1.8                  & 25.5\scriptsize±1.1 & 17.1\scriptsize±0.7                   & 18.7\scriptsize±0.7                  & 17.3\scriptsize±1.0\\
ER-DVC~\cite{gu2022not} (CVPR2022) & 56.2\scriptsize±3.1& 57.0\scriptsize±3.6                 & 61.2\scriptsize±3.5                 & 27.7\scriptsize±0.9                  & 28.5\scriptsize±1.1                  & 27.8\scriptsize±0.9 & 21.9\scriptsize±1.5                   & 23.8\scriptsize±1.2                  & 23.2\scriptsize±1.1 & 13.3\scriptsize±0.6                   & 14.6\scriptsize±0.8                  & 14.0\scriptsize±0.9\\
OCM~\cite{guo2022online} (ICML2022) & 56.7\scriptsize±3.0& 57.2\scriptsize±1.1                 & 65.3\scriptsize±4.2                 & 26.1\scriptsize±1.9                  & 26.5\scriptsize±2.3                  & 26.3\scriptsize±1.9 & 20.1\scriptsize±0.9                   & 21.9\scriptsize±0.9                  & 23.7\scriptsize±1.4 & 16.2\scriptsize±1.2                   & 17.9\scriptsize±1.1                  & 18.8\scriptsize±0.9\\
ER-ACE~\cite{caccia2021new} (ICLR2022)   & {60.8\scriptsize±2.7}& {62.0\scriptsize±2.9}                 & {65.3\scriptsize±2.6}                 & {31.5\scriptsize±1.4}                  & {33.2\scriptsize±1.1}                  & {33.3\scriptsize±1.5} & {27.8\scriptsize±1.8}                   & {30.7\scriptsize±0.8}                  & {29.8\scriptsize±1.2} & \underline{18.1\scriptsize±0.5}                   & \underline{21.7\scriptsize±0.5}                  & \underline{23.4\scriptsize±0.4}\\
OBC~\cite{chrysakisonline} (ICLR2023)   & 57.3\scriptsize±1.0& 61.3\scriptsize±1.0                 & 65.0\scriptsize±1.3                 & 30.0\scriptsize±1.1                  & 30.1\scriptsize±2.0                  & 30.9\scriptsize±1.9 & 22.4\scriptsize±0.7                   & 23.9\scriptsize±1.8                  & 25.9\scriptsize±1.5 & 15.2\scriptsize±0.7                   & 16.1\scriptsize±1.2                  & 17.9\scriptsize±1.1\\
OnPro~\cite{wei2023online} (ICCV2023)   & \underline{60.9\scriptsize±1.8}  & \underline{63.7\scriptsize±2.3}  & \underline{67.9\scriptsize±2.4}                 & 21.9\scriptsize±0.5  & 23.4\scriptsize±0.8  & 24.4\scriptsize±0.7 & 19.5\scriptsize±0.6  & 21.3\scriptsize±0.8  & 21.9\scriptsize±0.8 & 17.6\scriptsize±1.1                   & 19.9\scriptsize±0.8                  & 22.4\scriptsize±0.9\\
UER~\cite{lin2023uer} (ACMMM2023)   & {59.2\scriptsize±3.6}  & {60.8\scriptsize±2.2}  & {65.7\scriptsize±2.4}                 & \underline{32.4\scriptsize±0.8}  & \underline{33.9\scriptsize±1.2}  & \underline{34.3\scriptsize±1.8} & \underline{29.1\scriptsize±1.4}  & \underline{30.8\scriptsize±1.7}  & \underline{30.9\scriptsize±1.6} & 17.7\scriptsize±0.3                   & 20.0\scriptsize±0.9                  & 21.1\scriptsize±0.6\\
\hline
HPCR (Algorithm (\ref{alg:hpcr}))   & \textbf{65.2\scriptsize±3.1}& \textbf{66.0\scriptsize±2.7}                 & \textbf{71.0\scriptsize±2.6}                 & \textbf{40.1\scriptsize±1.4}                  & \textbf{40.6\scriptsize±1.0}                  & \textbf{42.4\scriptsize±1.4} & \textbf{36.0\scriptsize±1.3}                   & \textbf{37.9\scriptsize±1.1}                  & \textbf{38.2\scriptsize±1.2} & \textbf{26.6\scriptsize±0.8}                   & \textbf{28.3\scriptsize±0.9}                  & \textbf{29.4\scriptsize±0.6}\\
$\hookrightarrow$PCR   & 60.7\scriptsize±3.0& 60.3\scriptsize±2.8                 & 66.7\scriptsize±3.0                 & 33.7\scriptsize±1.3                  & 34.2\scriptsize±1.9                  & 34.7\scriptsize±1.9 & 30.6\scriptsize±0.9                   & 32.8\scriptsize±1.0                  & 33.0\scriptsize±1.1 & 21.3\scriptsize±0.5                   & 24.3\scriptsize±0.7                  & 25.5\scriptsize±0.5\\
\hline
\end{tabular}
\vspace{-4ex}
\end{table*}

\section{Performance Evaluation}
\label{sec:experiments}
\subsection{Experiment Setup}
\subsubsection{Datasets}
We conduct experiments on four image recognition datasets for evaluation. Split CIFAR10~\cite{krizhevsky2009learning} is split into 5 tasks, and each task contains 2 classes. Split CIFAR100~\cite{krizhevsky2009learning} as well as Split MiniImageNet~\cite{vinyals2016matching} are organized into 10 tasks, and each task is made up of samples from 10 classes. Split TinyImageNet~\cite{le2015tiny}, which contains 200 classes, is divided into 20 tasks, and each task contains 10 classes.

\subsubsection{Evaluation Metrics}
We need to measure the performance of the model for OCL. We first define $a_{i,j}(j<=i)$ as the accuracy evaluated on the held-out test samples of the $j$th task after the network has learned the training samples in the first $i$ tasks. Similar with~\cite{shim2021online}, we can further acquire the average accuracy rate
\begin{equation}
\label{eq:first}
 	\begin{split}
	 	A_i = \frac{1}{i}\sum_{j=1}^{i}a_{i,j},
 	\end{split}
\end{equation}
at the $i$th task. If there are $T$ tasks, we can get the final accuracy rate $A_T$ and the averaged anytime accuracy (AAA)
\begin{equation}
\label{eq:second}
 	\begin{split}
	 	AAA = \frac{1}{T}\sum_{t=1}^{T}A_{t},
 	\end{split}
\end{equation}
which measures the performance of the learning process~\cite{caccia2021new}.
Meanwhile, we denote the average forgetting rate as
\begin{equation}
\label{eq:first}
 	\begin{split}
F_i = \frac{1}{i-1}\sum_{j=1}^{i-1}f_{i,j},
 	\end{split}
\end{equation}
where $f_{k,j}= \mathop{\max}_{l\in\{1,2,...,k-1\}}(a_{l,j})-a_{k,j}$. And $F_T$ is equivalent to the final forgetting rate.

\subsubsection{Implementation Details}
The basic setting of the backbone is the same as the recent work~\cite{caccia2021new}. In detail, we take the Reduced ResNet18 (the number of filters is 20) as the feature extractor for all datasets. The classifier is NCM classifier for SCR and softmax classifier for other methods. During the training phase, the network, which is randomly initialized rather than pre-trained, is trained with the SGD optimizer, and the learning rate is set as 0.1. 
For all datasets, the classes are shuffled before division. And we set the memory buffer with different sizes for all datasets. The model receives 10 current samples from the data stream and 10 previous samples from the memory buffer at a time irrespective of the size of the memory. Moreover, we employ a combination of various augmentation techniques to get the augmented images. And the usage of data augmentation is fair for all methods.  Concerning hyperparameters of PCR, we select them on a validation set that is obtained by sampling 10\% of the training set. As for the testing phase, we set 256 as the batch size of validation.


\begin{figure*}[t]
\centering
    \begin{minipage}[t]{0.40\linewidth}
        \centering
        \includegraphics[scale=0.2]{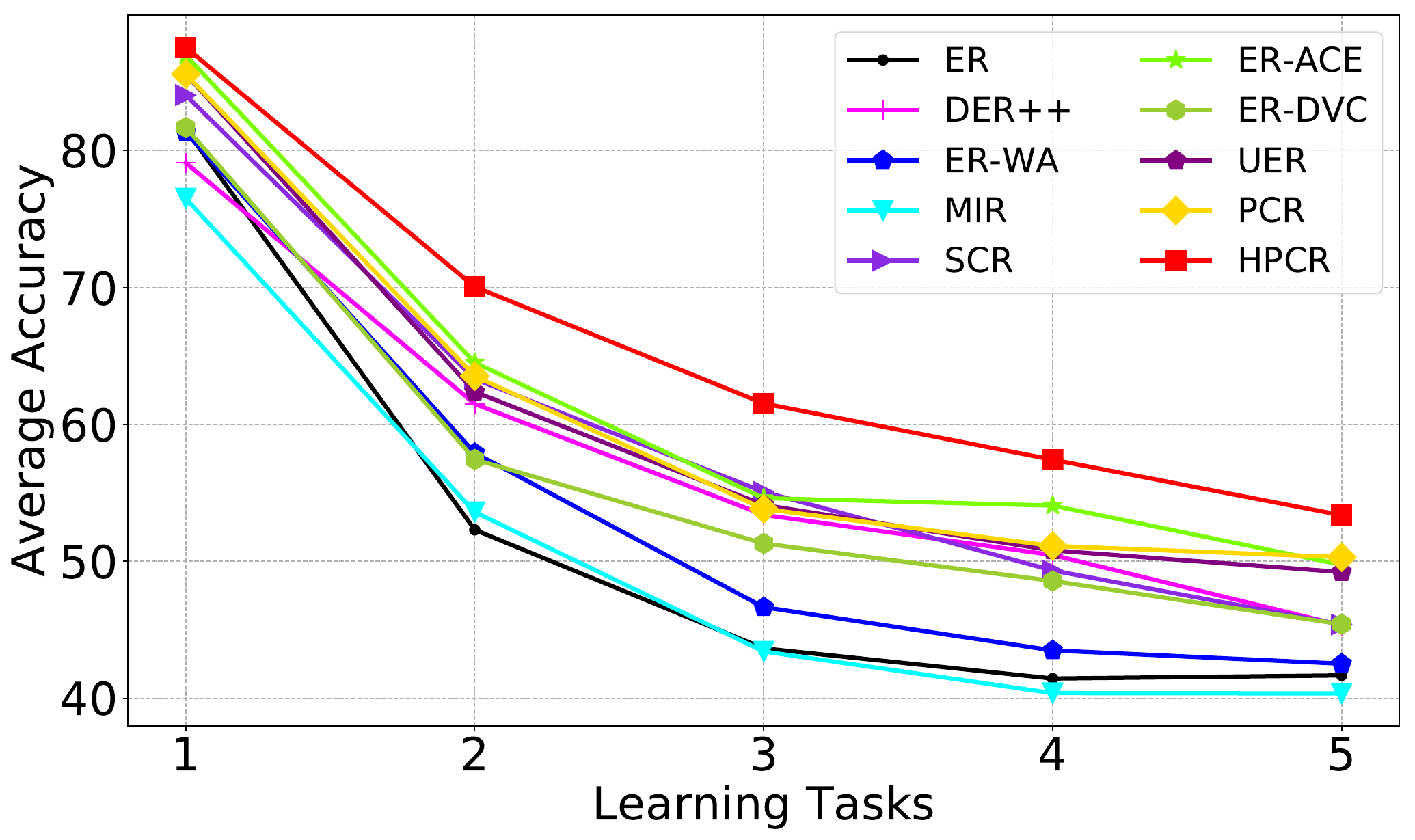}
        {(a) Split CIFAR10 (Buffer=200)}
    \end{minipage}
    \begin{minipage}[t]{0.40\linewidth}
        \centering
        \includegraphics[scale=0.2]{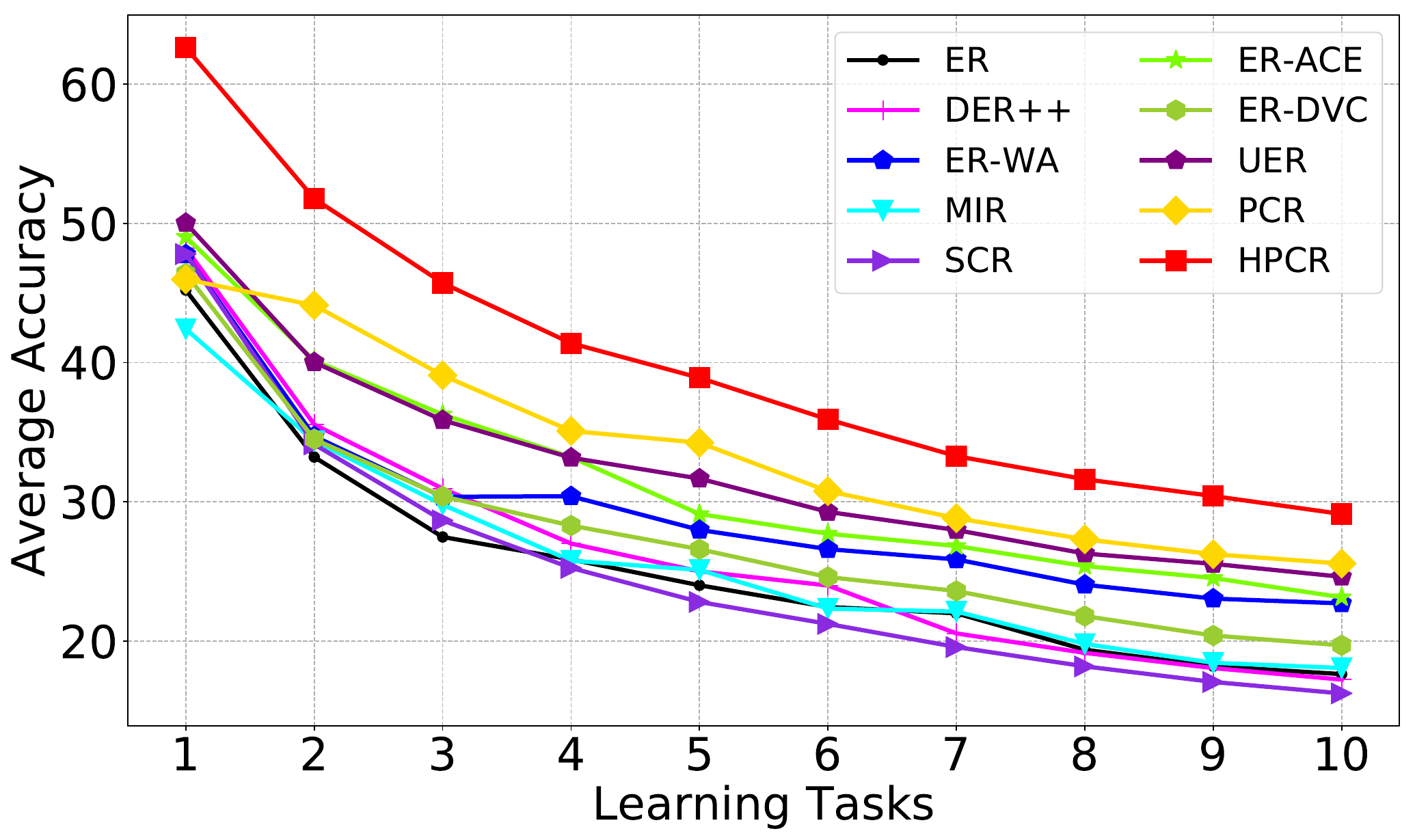}
        {(b) Split CIFAR100 (Buffer=1000)}
    \end{minipage}
    \begin{minipage}[t]{0.40\linewidth}
        \centering
        \includegraphics[scale=0.2]{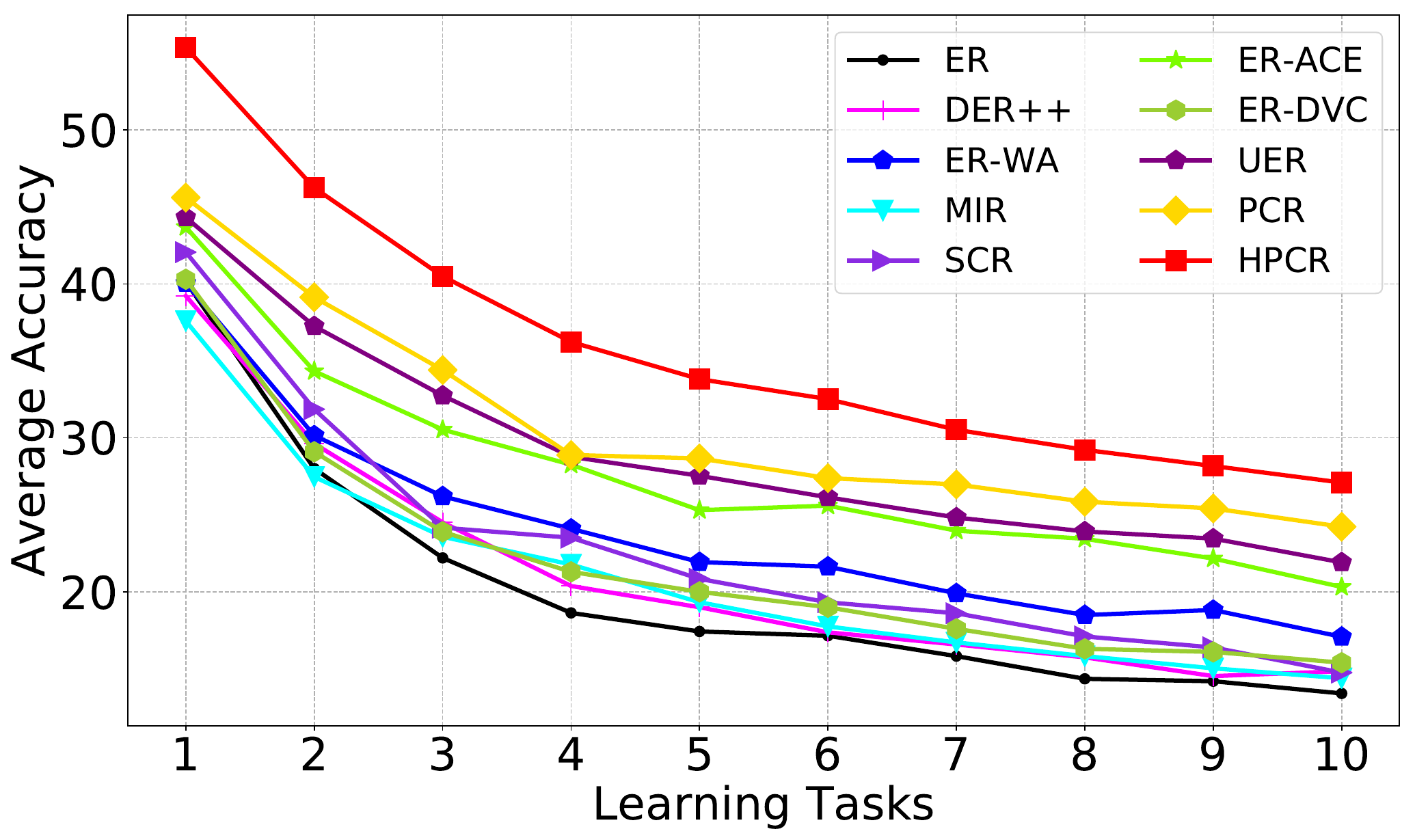}
        {(c) Split MiniImageNet (Buffer=1000)}
    \end{minipage}
    \begin{minipage}[t]{0.40\linewidth}
        \centering
        \includegraphics[scale=0.2]{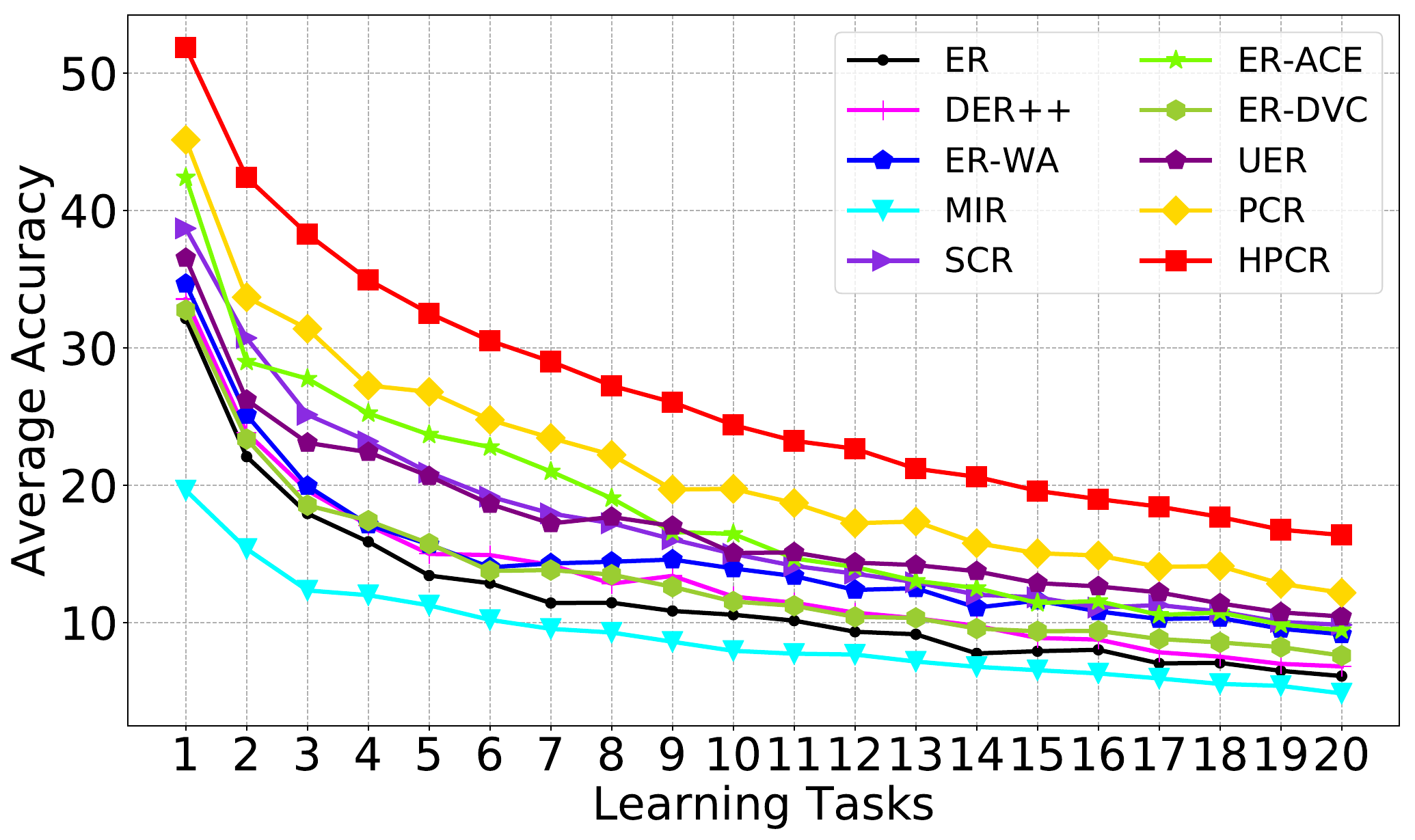}
        {(d) Split TinyImageNet (Buffer=2000)}
    \end{minipage}
\centering
\vspace{-1ex}
\caption{Average accuracy rate on observed learning tasks on all datasets. }
\label{learningstage1000}
\vspace{-2ex}
\end{figure*}

\begin{figure*}[t]
\centering
    \begin{minipage}[t]{0.4\linewidth}
        \centering
        \includegraphics[scale=0.2]{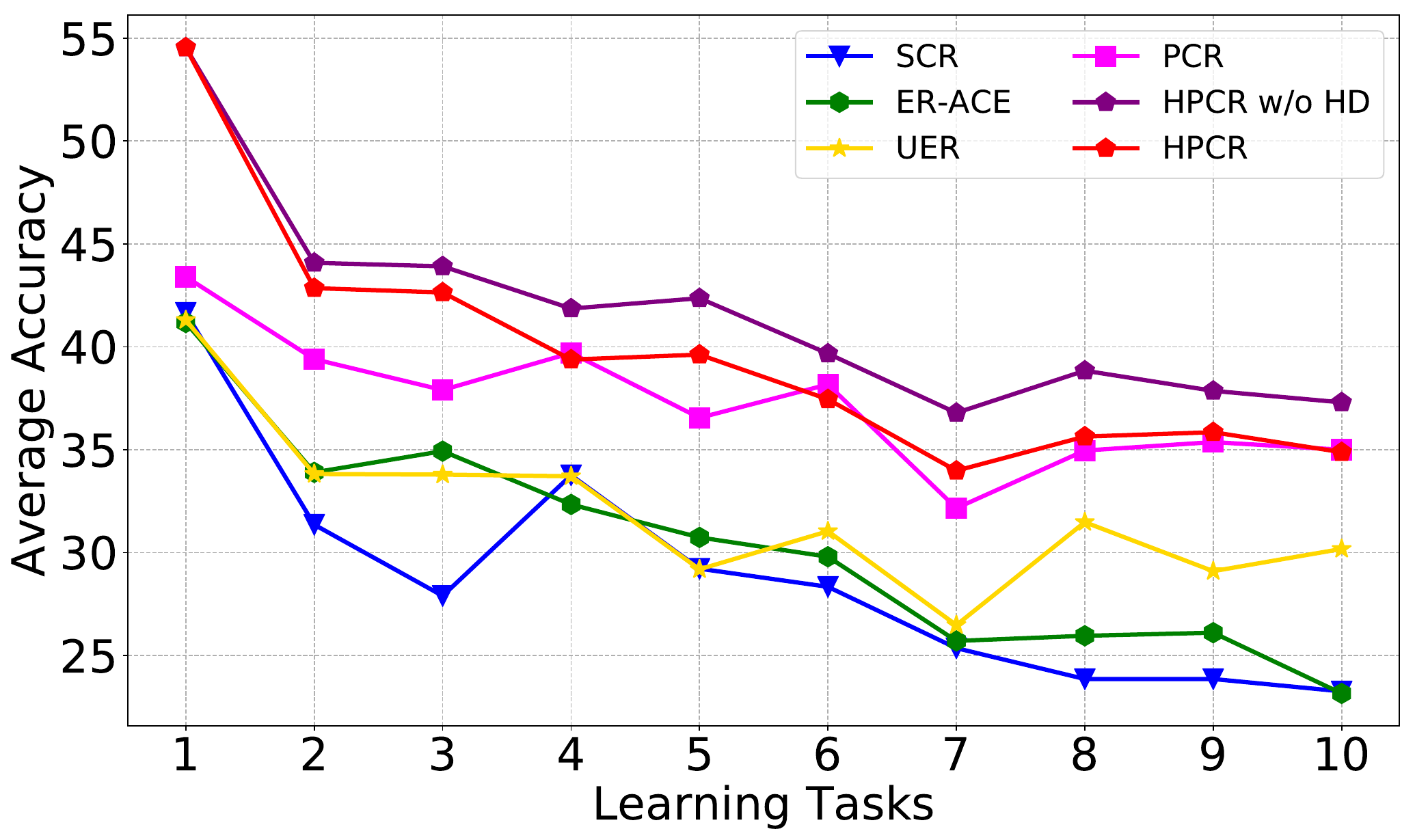}
        {(a) Performance on novel knowledge}
    \end{minipage}
    \begin{minipage}[t]{0.4\linewidth}
        \centering
        \includegraphics[scale=0.2]{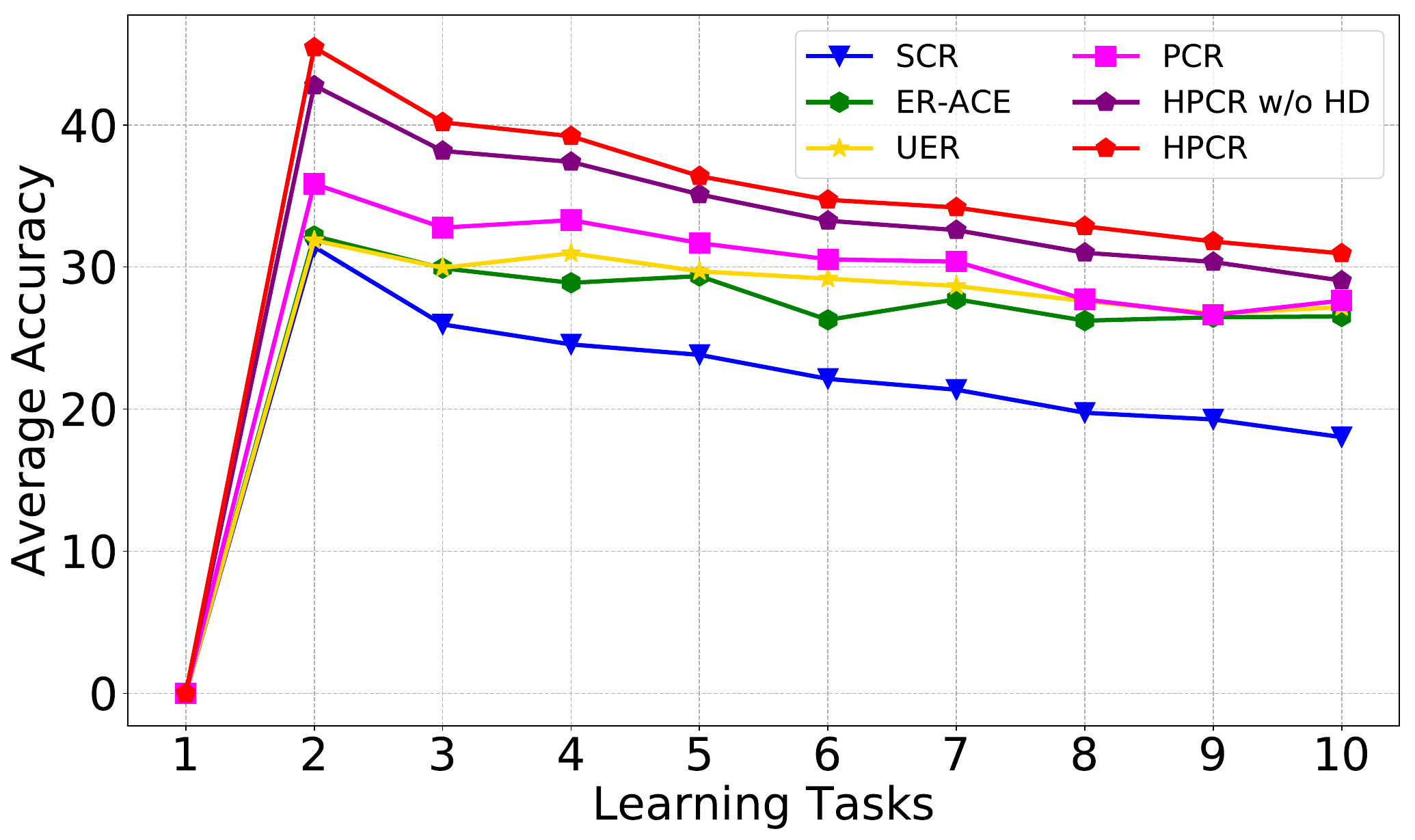}
        {(b) Performance on historical knowledge}
    \end{minipage}
\centering
\vspace{-1ex}
\caption{Average accuracy rate on observed learning tasks Split MiniImageNet while the buffer size is 5000. }
\label{learningstage5000}
\vspace{-2ex}
\end{figure*}

\begin{table*}[t]\small
\renewcommand\tabcolsep{2pt}
\centering
\caption{Final Forgetting Rate (↓) for four datasets.}
\vspace{-1ex}
\label{tableforget}
\begin{tabular}{l|ccc|ccc|ccc|ccc}
\hline
Datasets & \multicolumn{3}{c|}{Split CIFAR10 (\%)} & \multicolumn{3}{c|}{Split CIFAR100 (\%)}& \multicolumn{3}{c|}{Split MiniImageNet (\%)} & \multicolumn{3}{c}{Split TinyImageNet (\%)}\\ \hline
Buffer & \multicolumn{1}{c|}{100} & \multicolumn{1}{c|}{200} & \multicolumn{1}{c|}{500} & \multicolumn{1}{c|}{1000}    & \multicolumn{1}{c|}{2000} & \multicolumn{1}{c|}{5000} & \multicolumn{1}{c|}{1000} & \multicolumn{1}{c|}{2000}     &\multicolumn{1}{c|}{5000} & \multicolumn{1}{c|}{2000} & \multicolumn{1}{c|}{4000} & 10000     \\ \hline
SCR~\cite{mai2021supervised} (CVPR-W2021)   & 51.7\scriptsize±4.3 & 39.6\scriptsize±2.0                 & 27.7\scriptsize±2.9                 & 13.2\scriptsize±1.3                  & 12.9\scriptsize±1.5                  & 12.5\scriptsize±0.6 & 11.6\scriptsize±1.9                   & 11.5\scriptsize±1.4                  & 10.4\scriptsize±1.0 & 10.1\scriptsize±0.6                   & 8.7\scriptsize±0.6                  & 9.0\scriptsize±0.8\\
ER-ACE~\cite{caccia2021new} (ICLR2022)   & {18.7\scriptsize±1.1}& {17.1\scriptsize±2.5}                 & {13.8\scriptsize±1.7}                 & {10.5\scriptsize±0.8}                  & {9.8\scriptsize±0.5}                  & {7.9\scriptsize±1.5} & {9.8\scriptsize±1.1}                   & {7.3\scriptsize±1.4}                  & {6.2\scriptsize±0.9} & {9.2\scriptsize±0.9}                   & {7.5\scriptsize±0.4}                  & {6.4\scriptsize±0.7}\\
UER~\cite{lin2023uer} (ACMMM2023)   & {38.5\scriptsize±3.4}  & {30.7\scriptsize±3.6}  & {24.0\scriptsize±2.5}                 & {13.2\scriptsize±0.7}  & {9.8\scriptsize±1.0}  & {7.0\scriptsize±0.4} & {12.7\scriptsize±1.2}  & {8.4\scriptsize±0.7}  & {6.1\scriptsize±0.9} & 13.3\scriptsize±0.9                   & 9.0\scriptsize±1.0                  & 6.7\scriptsize±1.1\\
\hline

HPCR (Algorithm (\ref{alg:hpcr}))   & {28.5\scriptsize±1.2}& {20.6\scriptsize±2.4}                 & {15.5\scriptsize±1.8}                 & {15.7\scriptsize±0.7}                  & {11.1\scriptsize±0.5}                  & {9.3\scriptsize±0.8} & {15.4\scriptsize±0.6}                   & {11.1\scriptsize±0.7}                  & {8.6\scriptsize±0.4} & {20.5\scriptsize±0.8}                   & {13.6\scriptsize±1.1}                  & {9.2\scriptsize±0.7}\\
$\hookrightarrow$PCR   & 25.8\scriptsize±2.3& 19.4\scriptsize±3.3                 & 17.1\scriptsize±2.4                 & 15.5\scriptsize±0.9                  & 11.7\scriptsize±2.4                  & 9.5\scriptsize±0.8                  & 13.7\scriptsize±1.1 & 10.8\scriptsize±1.4                   & 9.5\scriptsize±1.5                  & 20.6\scriptsize±0.8 & 13.7\scriptsize±0.7                   & 10.2\scriptsize±0.6\\
\hline
\end{tabular}
\vspace{-4ex}
\end{table*}

\subsection{Overall Performance}
In this section, we conduct experiments to compare with various state-of-the-art baselines. The overall performance of PCR is advanced compared to all baselines. Moreover, HPCR obtains significantly improved performance on four datasets.

\textbf{Comparison on final accuracy.} HPCR has consistently achieved the best performance compared to all baseline methods. 
Table~\ref{tableaccuracy} shows the final accuracy rate for all datasets. All reported scores are the average score of 10 runs with 95\%
confidence interval. In a multitude of experimental scenarios, where each dataset comprises four memory buffers of varying sizes, HPCR consistently outshines all other methods. In particular, it exhibits even more remarkable performance compared to PCR. Specifically, HPCR shows a significant advancement, incorporating contrastive, temperature, and distillation components to further refine the PCR framework. For instance, on Split Cifar10, HPCR outperforms PCR by noteworthy margins of 2.9\%, 3.1\%, and 4.1\% for memory buffer sizes of 100, 200, and 500, respectively. These compelling results unequivocally highlight the crucial significance of the components introduced in this article, as they play a pivotal role in elevating the performance of PCR.

Moreover, we conduct a comprehensive comparative analysis between five methods, utilizing the experimental setup outlined in \cite{guo2022online}. In this particular study, the model is configured as ResNet18 with 64 filters, and 64 samples are retrieved from the memory buffer for each training batch. The training process employs the Adam optimizer with a learning rate 0.001. It is important to note that these experimental conditions differ from those employed in our own study. As presented in Table~\ref{moreaccuracy}, HPCR demonstrates remarkable superiority over SCR, OCM, OnPro, and PCR under different experimental settings. This evidence further reinforces the effectiveness and robustness of HPCR in comparison to existing approaches.

\begin{table*}[t]\small
\renewcommand\tabcolsep{2pt}
\centering
\caption{Final Accuracy Rate (higher is better) for ablation studies. ER acts as a baseline method.}
\vspace{-1ex}
\label{ablation}
\begin{tabular}{l|ccc|ccc|ccc|ccc}
\hline
Datasets & \multicolumn{3}{c|}{Split CIFAR10} & \multicolumn{3}{c|}{Split CIFAR100}& \multicolumn{3}{c|}{Split MiniImageNet}& \multicolumn{3}{c}{Split TinyImageNet}\\ \hline
Buffer & \multicolumn{1}{c|}{100} & \multicolumn{1}{c|}{200} & \multicolumn{1}{c|}{500} & \multicolumn{1}{c|}{1000} & \multicolumn{1}{c|}{2000} & \multicolumn{1}{c|}{5000}     & \multicolumn{1}{c|}{1000} & \multicolumn{1}{c|}{2000} & \multicolumn{1}{c|}{5000} & \multicolumn{1}{c|}{2000} & \multicolumn{1}{c|}{4000} & 10000     \\ \hline
ER & 33.8\scriptsize±3.2& 41.7\scriptsize±2.8                 & 46.0\scriptsize±3.5                 & 17.6\scriptsize±0.9                  & 19.7\scriptsize±1.6                  & 20.9\scriptsize±1.2 & 13.4\scriptsize±0.9                   & 16.5\scriptsize±0.9                  & 16.2\scriptsize±1.7                  & 6.1\scriptsize±0.5                  & 8.5\scriptsize±0.7                  & 8.9\scriptsize±0.6\\
Couple (Equation~(\ref{eq:coupleloss1})) & 38.4\scriptsize±3.0& 43.2\scriptsize±3.8                 & 49.1\scriptsize±3.9                 & 179\scriptsize±0.8                  & 19.7\scriptsize±0.8                  & 21.6\scriptsize±1.0 & 17.9\scriptsize±0.7                   & 19.9\scriptsize±0.8                  & 20.9\scriptsize±1.6                  & 5.2\scriptsize±0.4                  & 10.7\scriptsize±0.7                  & 12.8\scriptsize±0.5\\
Couple (Equation~(\ref{eq:coupleloss2})) & 34.1\scriptsize±2.3& 41.8\scriptsize±3.9                 & 46.3\scriptsize±2.8                 & 18.7\scriptsize±0.9                  & 20.7\scriptsize±0.8                  & 21.8\scriptsize±0.9 & 16.3\scriptsize±0.8                   & 18.7\scriptsize±0.7                  & 19.8\scriptsize±0.9                  & 6.2\scriptsize±0.3                  & 10.1\scriptsize±0.6                  & 11.0\scriptsize±0.8\\
PCR    & 45.4\scriptsize±1.3& 50.3\scriptsize±1.5                & 56.0\scriptsize±1.2                 & 25.6\scriptsize±0.6                &27.4\scriptsize±0.6                 & 29.3\scriptsize±1.1 & 24.2\scriptsize±0.9                   & 27.2\scriptsize±1.2                  & 28.4\scriptsize±0.9                   & 12.2\scriptsize±0.9                   & 17.4\scriptsize±0.7                  & 19.6\scriptsize±0.8\\\hline

PCR+HC ($\rm PCR_C$)    & 41.9\scriptsize±2.0& 48.3\scriptsize±2.4                & 55.8\scriptsize±1.2                 & 24.1\scriptsize±0.5           & 25.9\scriptsize±0.5          & 27.3\scriptsize±0.7 & 21.8\scriptsize±0.8                   & 24.5\scriptsize±0.9                  & 26.3\scriptsize±1.0                   & {11.1\scriptsize±1.2}                   & {13.6\scriptsize±0.7}                  & {15.3\scriptsize±1.2}\\

PCR+HC+HT ($\rm PCR_{CT}$)    & {47.4\scriptsize±2.2}& {51.3\scriptsize±1.6}                & {57.7\scriptsize±1.1}                 & {27.2\scriptsize±0.5}                &{29.3\scriptsize±0.7}                 & {31.6\scriptsize±0.9} & {25.6\scriptsize±1.0}                   & {28.5\scriptsize±0.5}                  & {29.8\scriptsize±0.7} & {14.8\scriptsize±0.6}                   & {18.8\scriptsize±0.5}                  & {20.8\scriptsize±0.5}\\
$\rm PCR_{CT}$+PCD    & 48.0\scriptsize±1.9& 52.4\scriptsize±1.2                & 59.1\scriptsize±1.2                 & 28.6\scriptsize±0.8           & 30.1\scriptsize±0.7          & 33.0\scriptsize±0.5 & 26.6\scriptsize±0.5                   & 29.3\scriptsize±0.6                  & 30.9\scriptsize±0.7                   & 15.6\scriptsize±0.3                  & 19.2\scriptsize±0.5                  & 21.8\scriptsize±0.6\\
$\rm PCR_{CT}$+PCD+SCD (HPCR)    & \textbf{48.3\scriptsize±1.5}& \textbf{53.4\scriptsize±1.4}                & \textbf{60.1\scriptsize±1.1}                 & \textbf{29.1\scriptsize±0.7}                &\textbf{30.7\scriptsize±0.5}                 & \textbf{33.7\scriptsize±0.6} & \textbf{27.1\scriptsize±0.6}                   & \textbf{29.9\scriptsize±0.7}                  & \textbf{31.3\scriptsize±0.7} & \textbf{16.4\scriptsize±0.3}                   & \textbf{19.5\scriptsize±0.8}                  & \textbf{22.1\scriptsize±0.5}\\
\hline
\end{tabular}
\vspace{-3ex}
\end{table*}

\textbf{Comparison on learning process.} In the meantime, HPCR performs the best overall throughout the entire learning process. Specifically, we report the averaged anytime accuracy for all datasets in Table~\ref{tableaaa}. The experimental results show that HPCR consistently outperforms other baselines during the whole learning process. For a more detailed comparison, we also report the accuracy of each learning task in Figure~\ref{learningstage1000}, which can further validate the overall performance of the methods. The results show that the performance of PCR in the first few tasks does not outperform other baselines. However, its improvement becomes more and more visible as the number of tasks increases, proving its power to overcome CF. For instance, PCR does not perform best in the first task, but it demonstrates outstanding advantages in the remaining tasks on Split CIFAR100. Fortunately, this issue has been resolved in HPCR. The comparison of the red and orange lines shows that HPCR still performs well in the first few tasks. Therefore, our approaches have a stronger ability to resist forgetting.

\textbf{Comparison on knowledge balance.} Finally, HPCR also has significant advantages in balancing historical and novel knowledge. Before this, we calculate the final forgetting rate of the model when using different methods. As reported in Table~\ref{tableforget}, the existing methods, especially ER-ACE, have shown outstanding performance in this evaluation metric.
This is because ER-ACE tends to learn less, making it easily forget less. To verify it, we record the accuracy of novel and historical knowledge in each task for part of effective methods on Split MiniImageNet. As demonstrated in Figure~\ref{learningstage5000}, as ER-ACE learns less novel knowledge, its performance on the final forgetting rate can be better.

Actually, we should not only focus on the model's ability to retain historical knowledge, but also ensure the model's ability to quickly learn novel knowledge. Although SCR, ER-ACE, and UER can improve the anti-forgetting ability of the model, they tend to limit the generalization ability of the model. As the historical knowledge is consolidated, the learning performance of the model on novel knowledge becomes very poor. Different from existing studies, our model can not only effectively alleviate the phenomenon of CF, but also reduce the decline of the model at the generalization level as much as possible. Compared with HPCR (red line) and PCR (orange line), it is not difficult to find that there is a significant improvement in the memory ability of old knowledge.

\begin{figure}[t]
\centering
\includegraphics[scale=0.18]{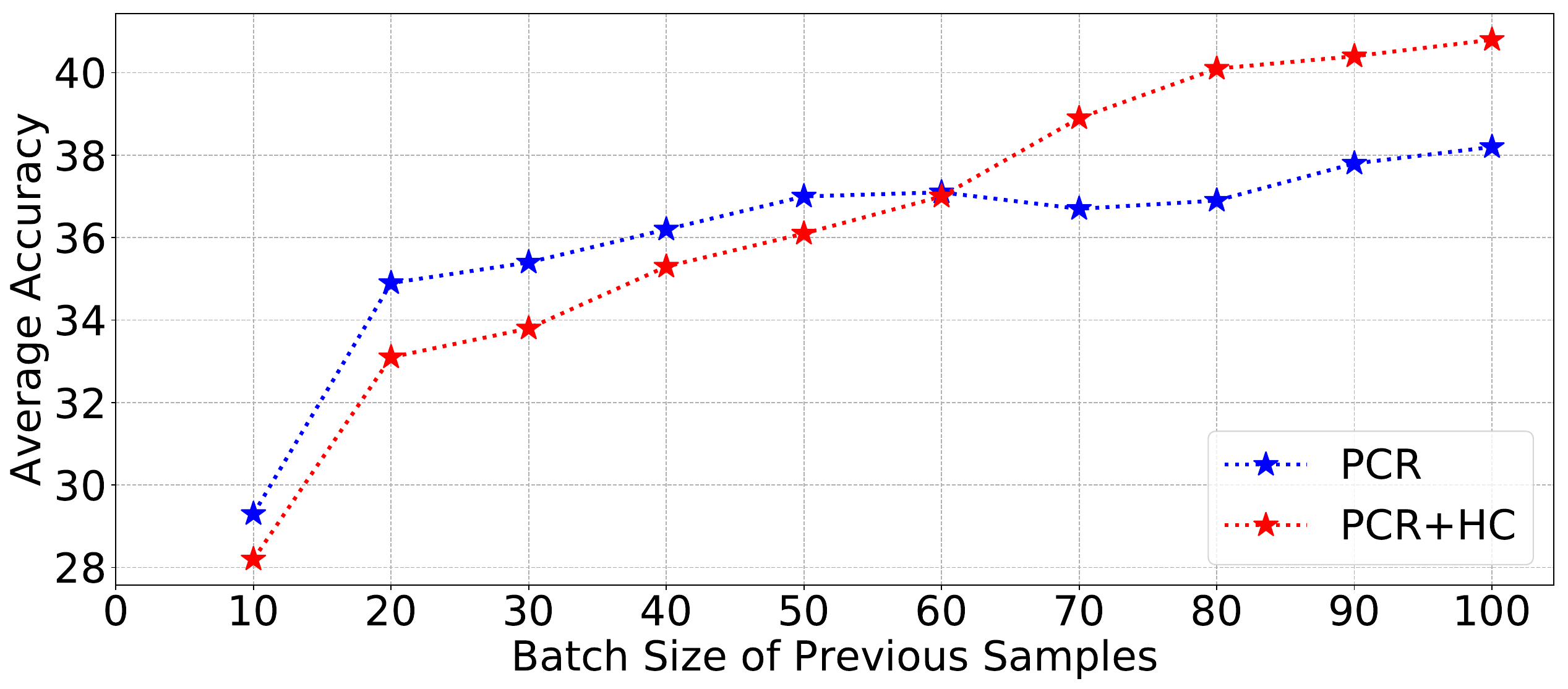}
\vspace{-2ex}
\caption{Final Accuracy Rate on Split CIFAR100 (buffer size=5000) with different batch sizes of previous samples.}
\label{fig:ablation_hc}%
\vspace{-2ex}
\end{figure}

\begin{figure}[t]
\centering
\includegraphics[scale=0.18]{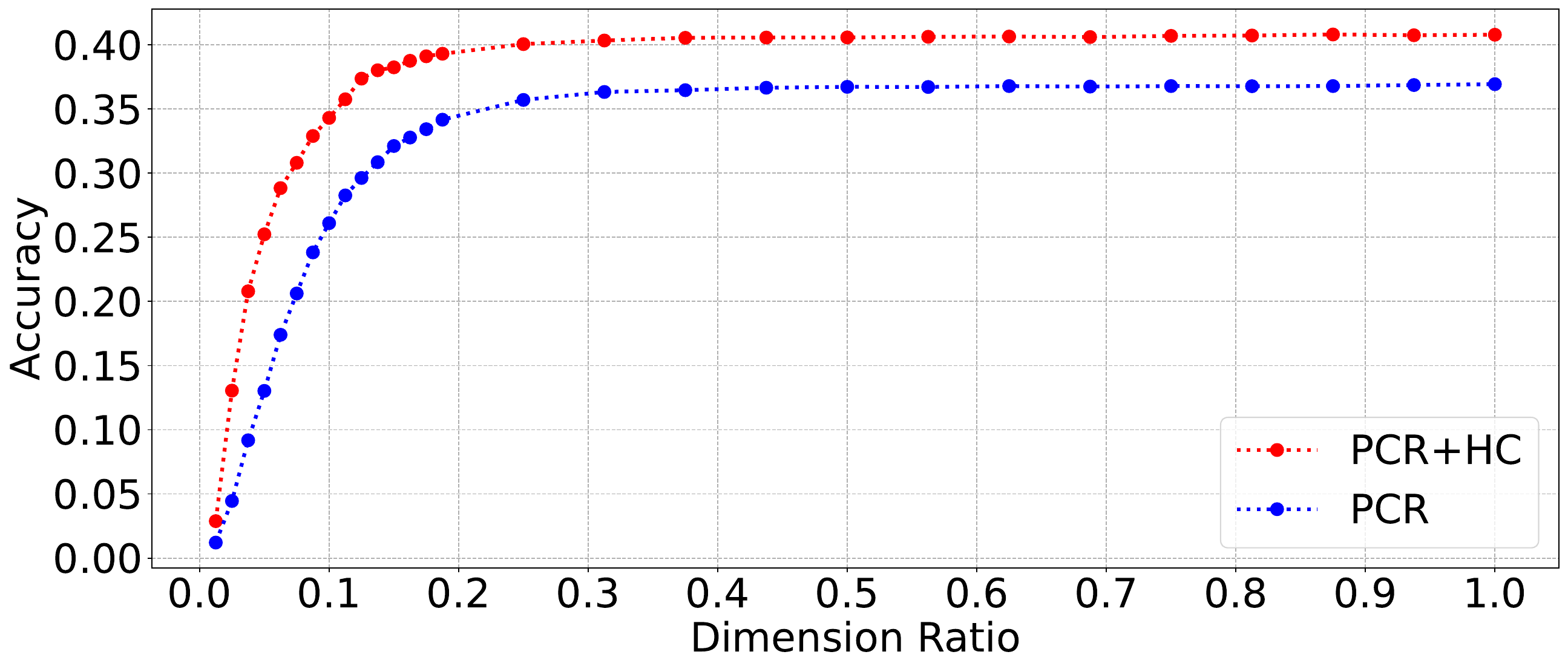}
\vspace{-2ex}
\caption{Final Accuracy Rate on Split CIFAR100 (buffer size=5000) with different ratios of feature dimension.}
\label{fig:ablation_hc_pca}%
\vspace{-2ex}
\end{figure}

\begin{table}[t]
\renewcommand\tabcolsep{2.5pt}
\centering
\caption{Analysis results on CIFAR10 (buffer size=100) with and without temperature component.}
\vspace{-2ex}
\label{table:ablation_ht}
\begin{tabular}{l|lllll}
\hline
\makecell[l]{Different\\Setting} & \makecell[l]{Final\\Accuracy} & \makecell[l]{Cumulative\\gradient} & \makecell[l]{Loss} & \makecell[l]{Difference \\(all classes)} & \makecell[l]{Difference \\(new classes)} \\ \hline
w/o HT  & 42.83  & 88.27/-95.39    & 1.8439        & 0.0042     & 0.0056    \\
with HT & 47.66 & 64.28/-70.40    & 1.6710        & 0.0025     & 0.0040     \\ \hline
\end{tabular}
\vspace{-2ex}
\end{table}

\begin{figure}[t]
\centering
\includegraphics[scale=0.6]{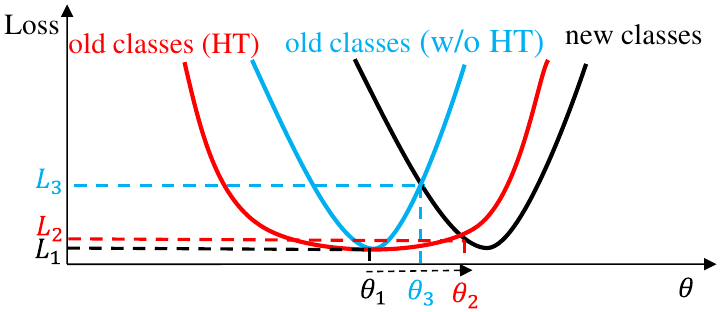}
\vspace{-2ex}
\caption{Illustration of a steep region not using HT and a flat region using HT.}
\label{fig:ablation_ht}%
\vspace{-2ex}
\end{figure}

\begin{table}[t]
\renewcommand\tabcolsep{2.5pt}
\centering
\caption{Accuracy Rate (higher is better) of HPCR on Split CIFAR100 with different $\tau_{min}$ and $\tau_{max}$ when the cycle length is 500. }
\vspace{-2ex}
\label{table:tau_step}
\begin{tabular}{l|lllll}
\hline
\diagbox{$\tau_{min}$}{$\tau_{max}$} & 0.12       & 0.14       & 0.16                & 0.18       & 0.2        \\ \hline
0.11  & 25.4, 34.9 & 25.0, 36.9 & 26.2, 36.6          & 26.3, 36.6 & 25.9, 36.7 \\
0.09  & 25.5, 35.8 & 26.0, 36.7 & 26.1, 36.9          & 26.5, 37.6 & 26.1, 37.8 \\
0.07  & 26.6, 36.8 & 26.3, 37.1 & 26.4, 37.6          & 26.2, 37.3 & 27.0, 38.1 \\
0.05  & 26.6, 37.1 & 26.7, 37.7 & \textbf{27.7, 38.5} & 26.9, 37.9 & 26.5, 37.6 \\ \hline
\end{tabular}
\vspace{-2ex}
\end{table}

\begin{table}[t]
\renewcommand\tabcolsep{2.5pt}
\centering
\caption{Accuracy Rate (higher is better) of HPCR on Split CIFAR100 with different cycle lengths $S$.}
\vspace{-2ex}
\label{table:tau_bound}
\begin{tabular}{l|llllll}
\hline
$S$ & 50   & 100  & 125  & 250  & 500  & 1000 \\ \hline
Final Accuracy   & 24.7 & 25.7 & 25.8 & 26.7 & \textbf{27.7} & 23.2 \\
Averaged Accuracy   & 34.6 & 36.5 & 36.8 & 37.8 & \textbf{38.5} & 32.4 \\ \hline
\end{tabular}
\vspace{-2ex}
\end{table}

\begin{table}[!h]
\renewcommand\tabcolsep{2.5pt}
\centering
\caption{Final Accuracy Rate (higher is better) of each task on Split CIFAR10 using distillation as DER or PCD.}
\vspace{-2ex}
\label{table:ablation_hd}
\begin{tabular}{l|llllll}
\hline
Task     & 1               & 2              & 3              & 4      & 5      & Average         \\ \hline
DER~\cite{buzzega2020dark} & 0.3480           & 0.2205         & 0.4320          & \textbf{0.5315} & \textbf{0.7500}   & 0.4564          \\
PCD (ours)      & \textbf{0.3695} & \textbf{0.2270} & \textbf{0.488} & 0.4960  & 0.7405 & \textbf{0.4642} \\ \hline
\end{tabular}
\vspace{-2ex}
\end{table}

\begin{figure}[t]
\centering
\includegraphics[scale=0.18]{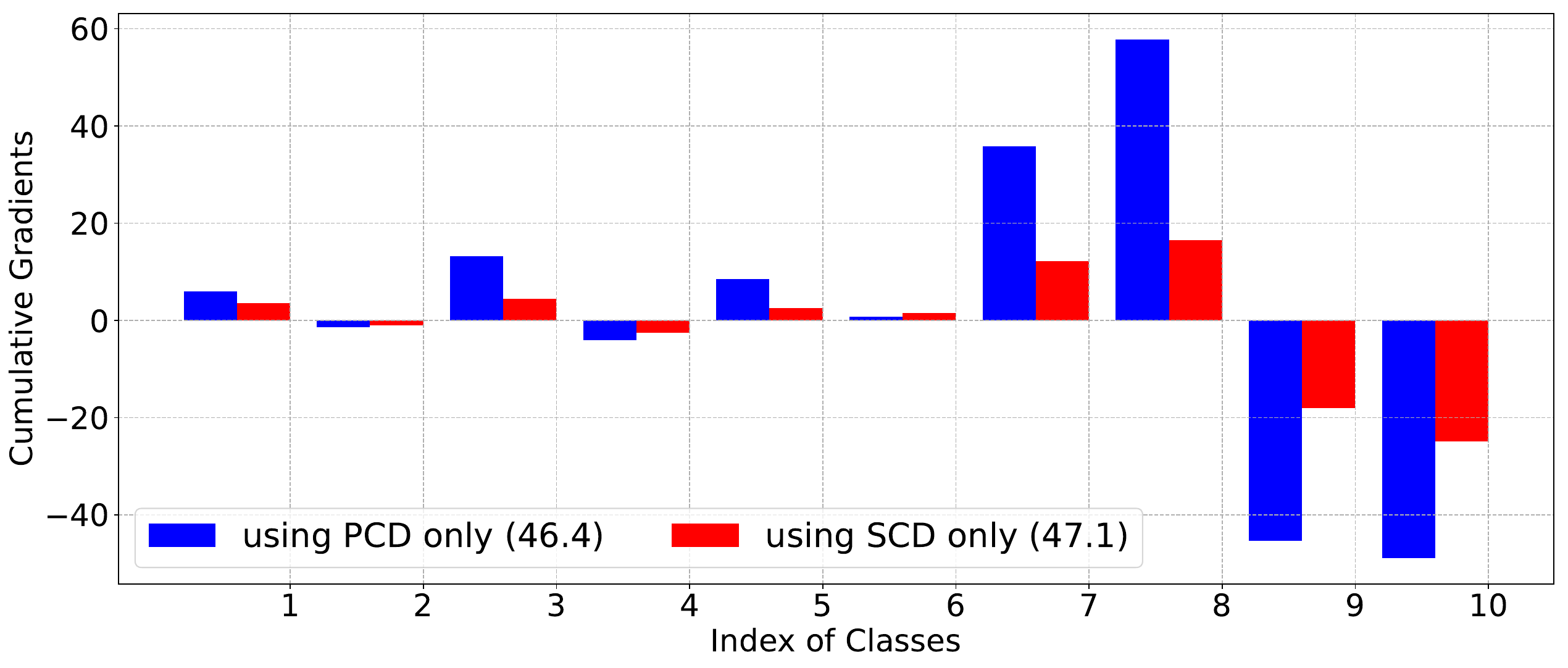}
\vspace{-2ex}
\caption{Cumulative gradient for all proxies using different distillation loss functions when the model learns new classes (9/10) on Split CIFAR10.}
\label{fig:ablation_hd}%
\vspace{-2ex}
\end{figure}

\begin{figure}[t]
\centering
\includegraphics[scale=0.22]{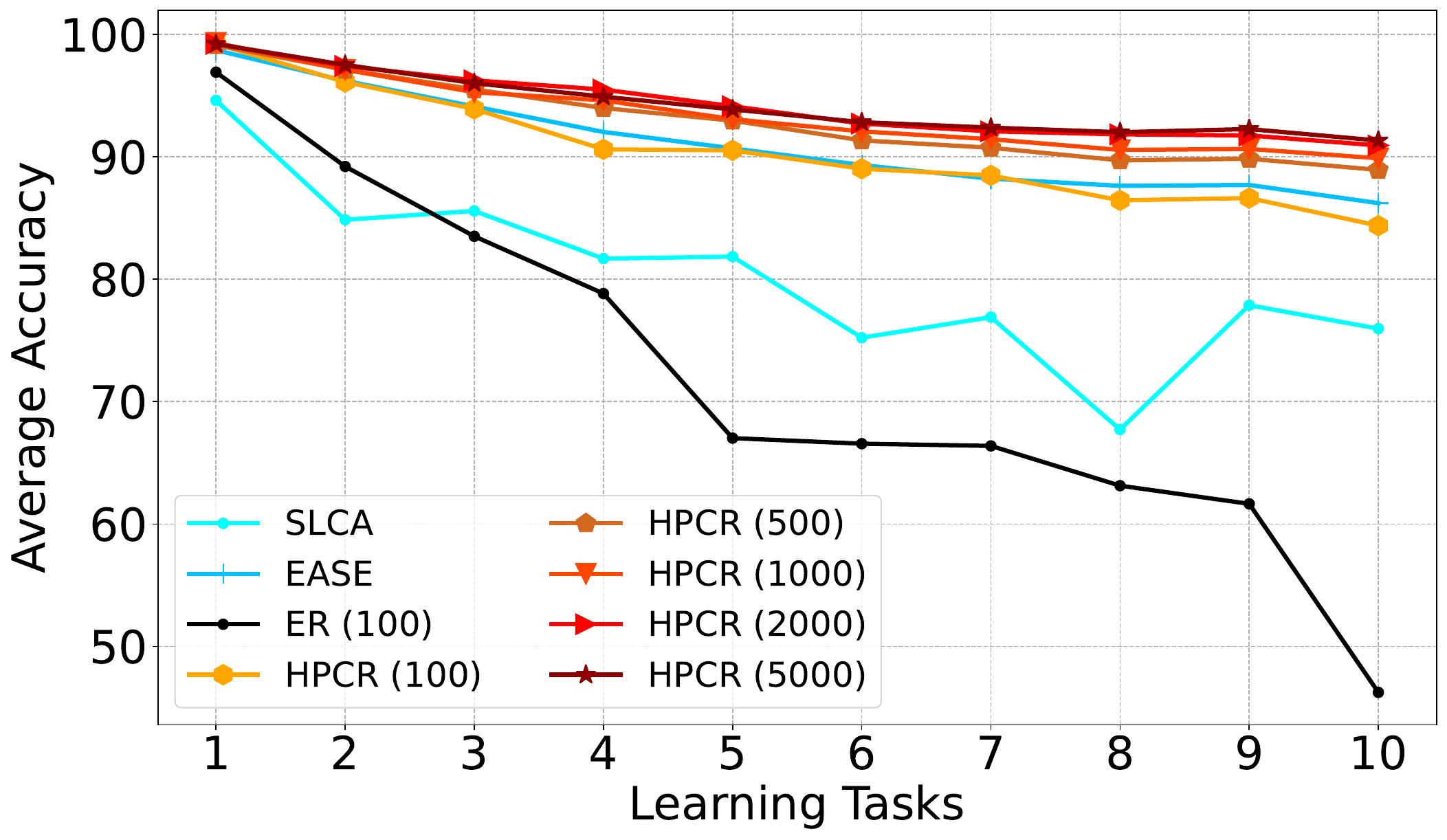}
\vspace{-2ex}
\caption{Average accuracy Rate on Split CIFAR100 with different methods.}
\label{fig:pretrained}%
\vspace{-2ex}
\end{figure}

\begin{figure}[!h]
\centering
\subfigure[ER (previous samples=10)]{
    \begin{minipage}[t]{0.46\linewidth}
        \centering
        \includegraphics[scale=0.13]{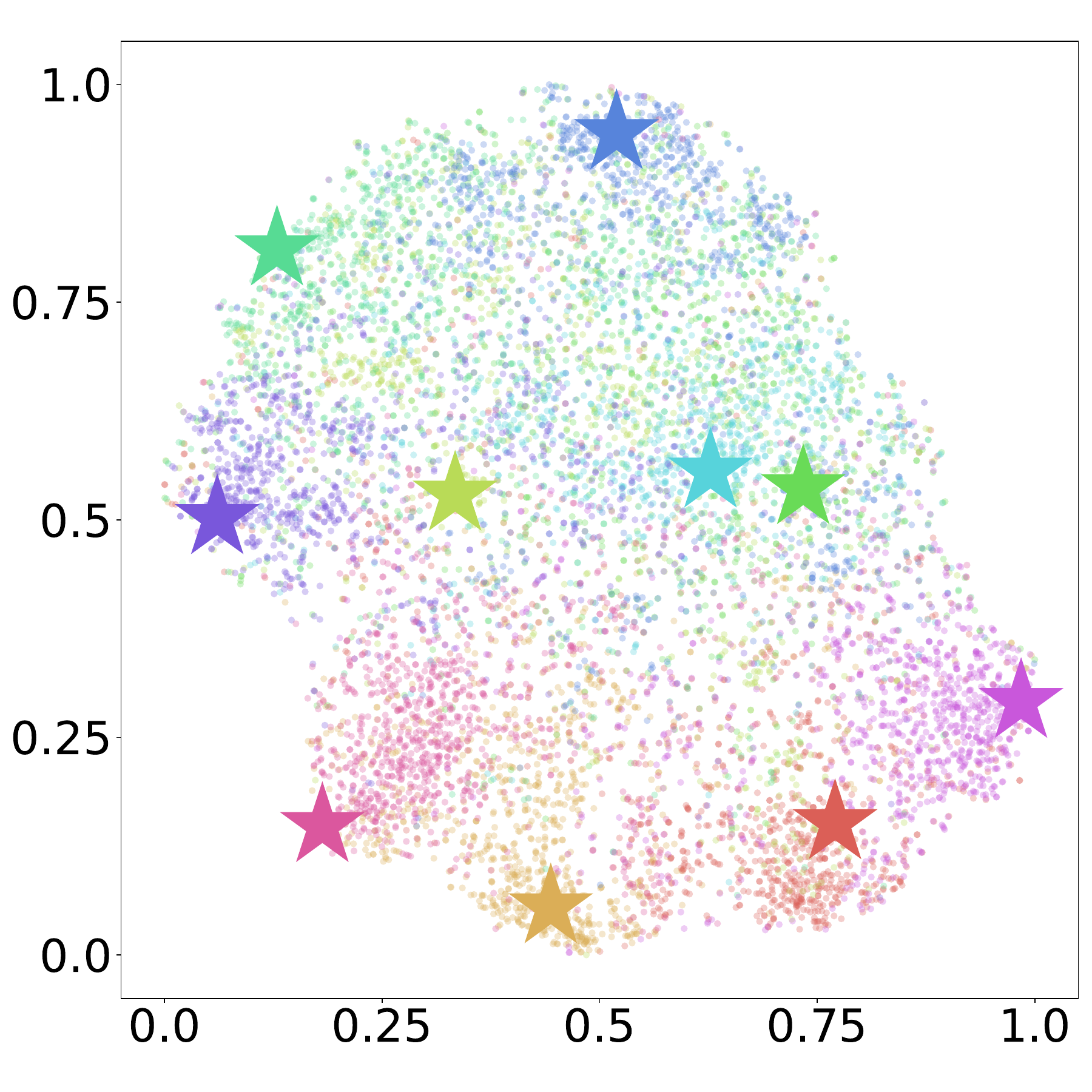}
    \end{minipage}
}
\subfigure[ER (previous samples=100)]{
    \begin{minipage}[t]{0.46\linewidth}
        \centering
        \includegraphics[scale=0.13]{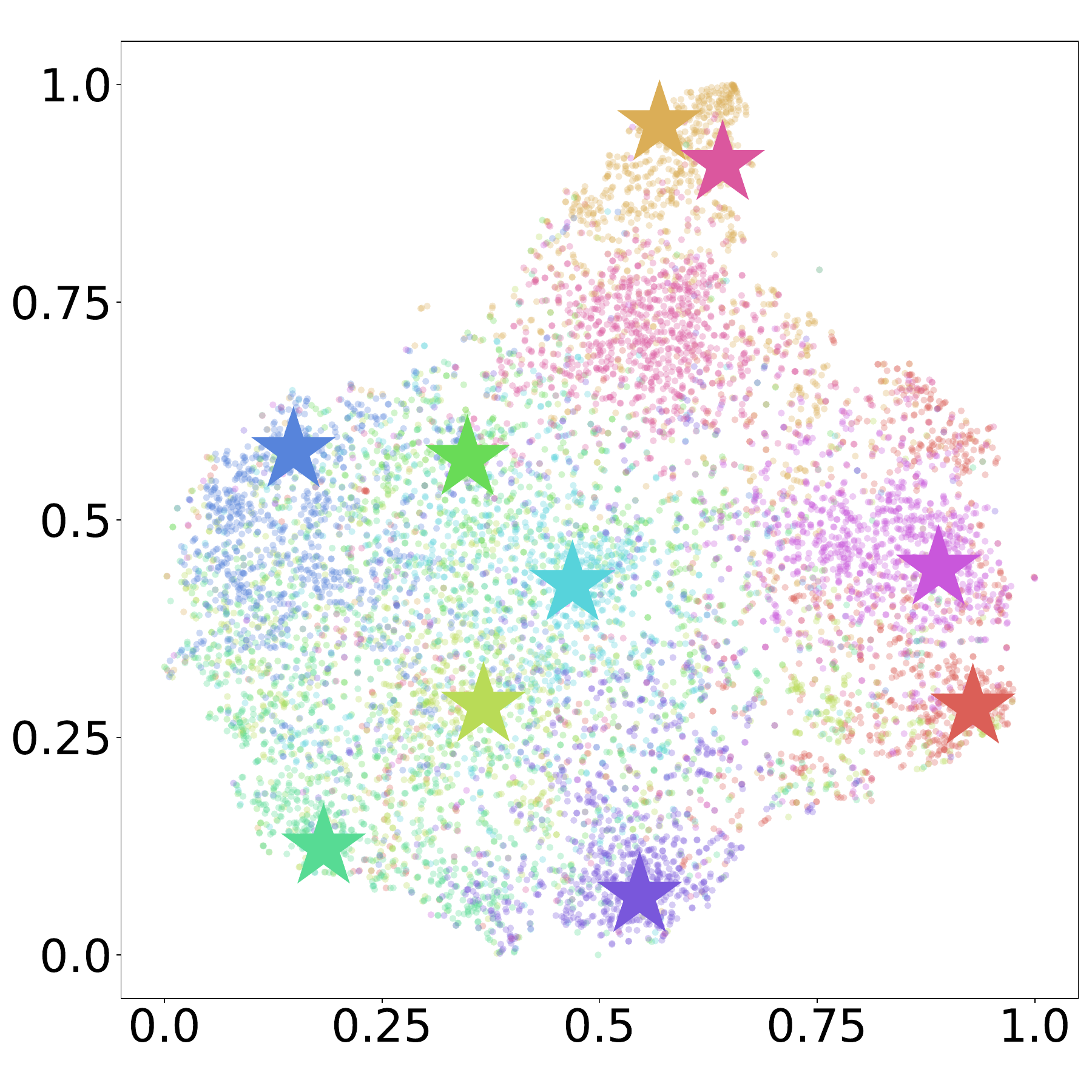}
    \end{minipage}
}
\centering
\subfigure[HPCR (previous samples=10)]{
    \begin{minipage}[t]{0.46\linewidth}
        \centering
        \includegraphics[scale=0.13]{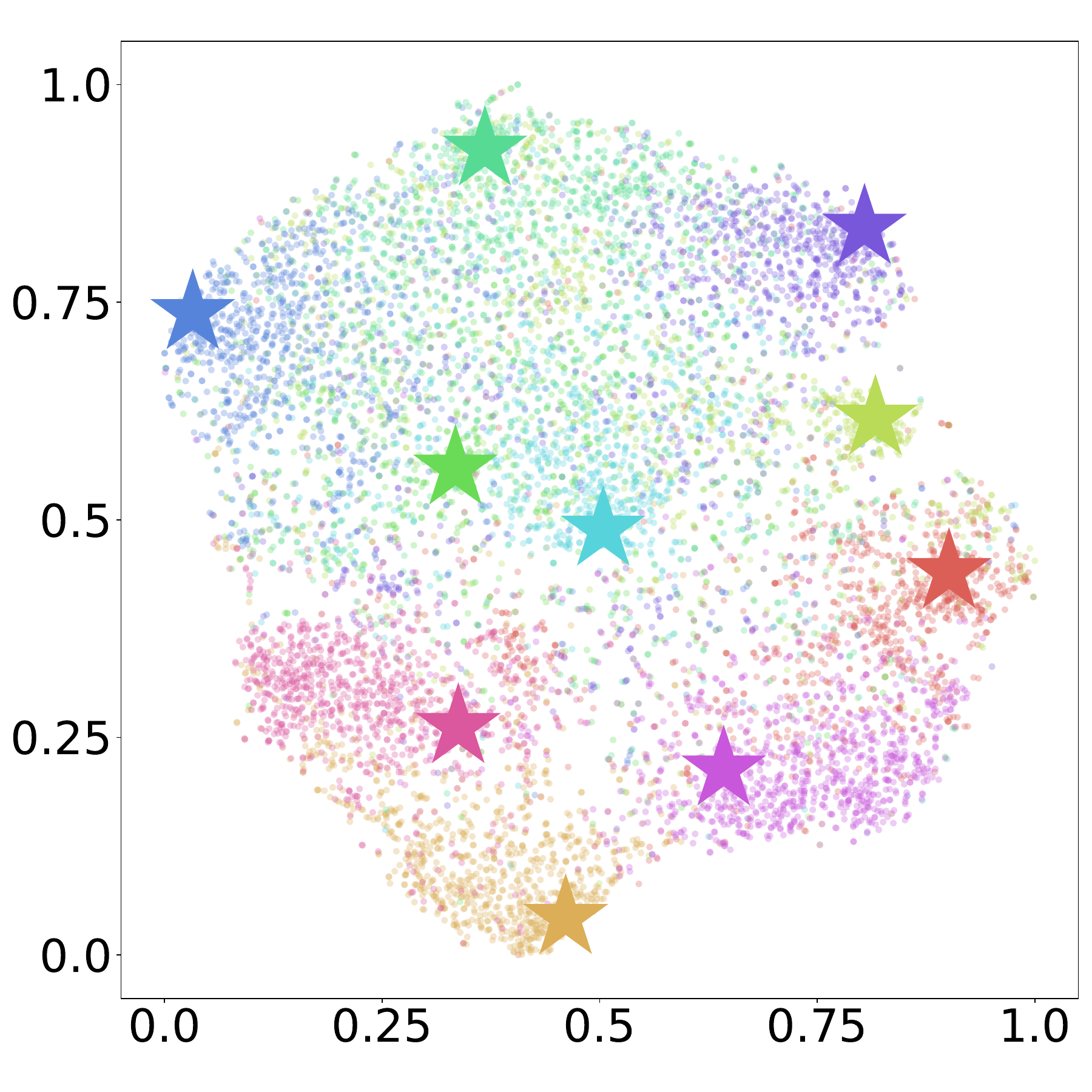}
    \end{minipage}
}
\subfigure[HPCR (previous samples=100)]{
    \begin{minipage}[t]{0.46\linewidth}
        \centering
        \includegraphics[scale=0.13]{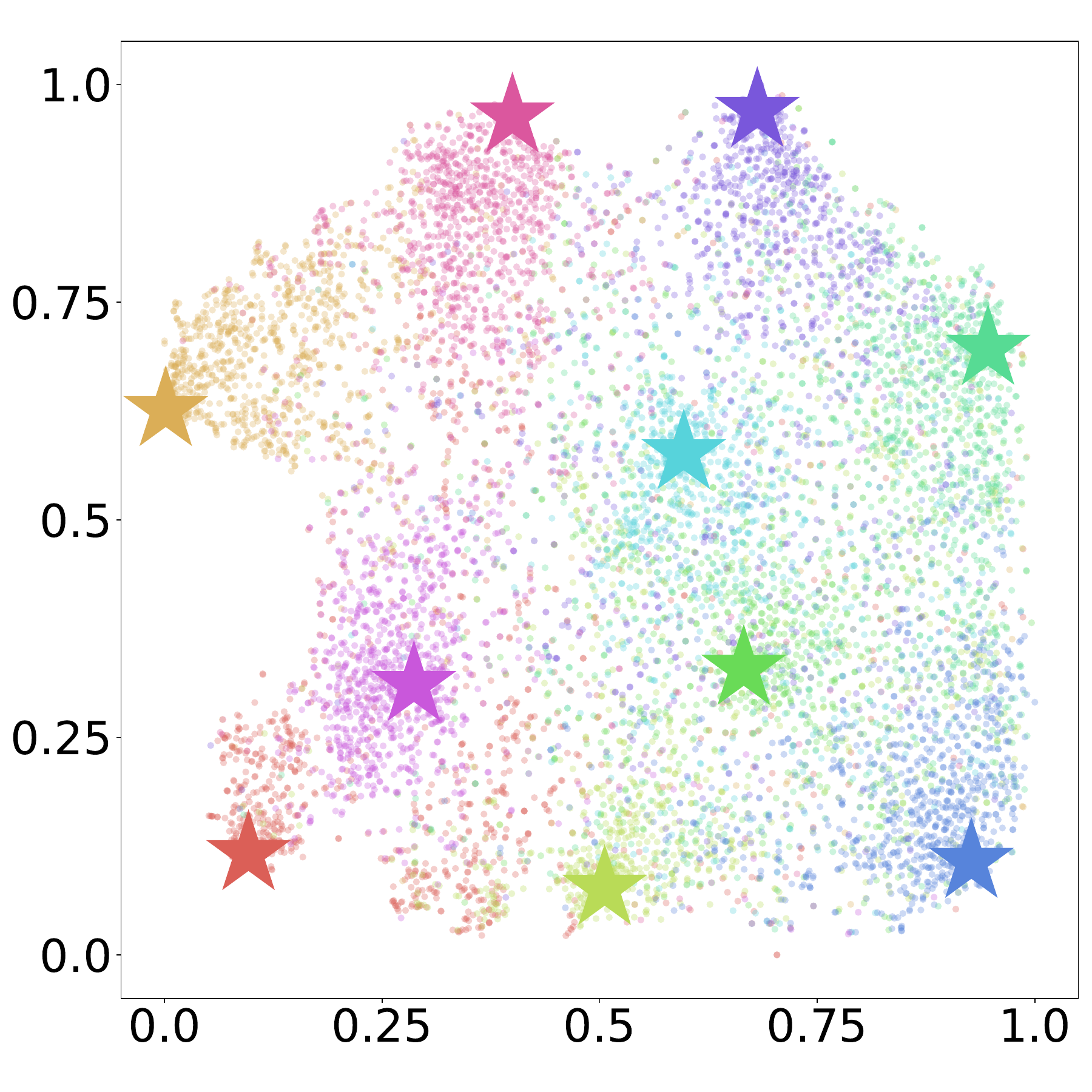}
    \end{minipage}
}
\vspace{-1ex}
\caption{2D t-SNE visualization of feature embeddings for the testing samples at the end of the training on Split CIFAR10 (buffer size=500).} 
\label{fig:tsne}
\vspace{-3ex}
\end{figure}

\begin{table}[t]
\renewcommand\tabcolsep{2pt}
\centering
\caption{The computation and memory budget of different methods.}
\vspace{-2ex}
\label{table:other}
\begin{tabular}{l|ccccccc}
\hline
Method       & ER      & ER-ACE & OCM & OnPro & UER & PCR & HPCR\\ \hline
Computation ($C_\mathcal{D}$)       & 1      & 1 & 16 & 16 & 1 & 1 & 1\\
Memory (Model)       & 1      & 1 & 2 & 1 & 1 & 1 & 1\\
Memory (Data)       & 1      & 1 & 1 & 1 & 1 & 1 & Table~\ref{table:memory}\\
\hline
\end{tabular}
\vspace{-2ex}
\end{table}

\begin{table}[t]
\renewcommand\tabcolsep{2pt}
\centering
\caption{The memory (data) on different datasets for HPCR.}
\vspace{-2ex}
\label{table:memory}
\begin{tabular}{l|cccc}
\hline
Dataset       & CIFAR10      &CIFAR100  & MiniImageNet & TinyImageNet\\ \hline
Sample Size       & 32×32×3      & 32×32×3 & 84×84×3 & 64×64×3\\
Feature Size        & 160      & 160 & 640 & 640\\
Logits Size       & 10      & 100 & 100 & 200\\
Memory (Data)       & 1.06      & 1.08 & 1.03 & 1.07\\
Buffer Size     & 100      & 1000 & 1000 & 2000\\
Accuracy (Control)       & 47.7\scriptsize±1.5      & 27.4\scriptsize±0.5 & 26.5\scriptsize±0.6 & 15.2\scriptsize±0.4\\
Accuracy (HPCR)       & 48.3\scriptsize±1.5      & 29.1\scriptsize±0.7 & 27.1\scriptsize±0.6 & 16.4\scriptsize±0.3\\
\hline
\end{tabular}
\vspace{-4ex}
\end{table}

\subsection{Ablation Study}
In this section, we decompose HPCR into several components and demonstrate their functions. We conduct experiments with different settings and record their final accuracy rate in Table~\ref{ablation}. In addition to ER, we also include the two combination methods denoted as Equation (\ref{eq:coupleloss1}) and Equation (\ref{eq:coupleloss2}). The experimental results show that they are not significant since they do not solve the problem of forgetting.

\textbf{Contrastive component} (HC) conditionally incorporates anchor-to-sample pairs to PCR. Although it provides more knowledge about the relationships of samples, its performance is limited by a small batch size of training samples, as the ``PCR+HC'' indicated in Table~\ref{ablation}. Meanwhile, we conduct experiments on Split CIFAR100 using PCR with and without HC. The results revealed in Figure~\ref{fig:ablation_hc} show that the anchor-to-sample pairs can reduce the performance of PCR when the batch size of the training samples is small ($<60$), and vice versa. Hence, the usage of anchor-to-sample pairs should be controlled by a stage function denoted as Equation (\ref{eq:step_function}). To further explore HC, we conduct principal component analysis (PCA)~\cite{wold1987principal} on the features $\bm{z}$ extracted by the model. As demonstrated in Figure~\ref{fig:ablation_hc_pca}, we report the final accuracy rate on Split CIFAR100 with different ratios of feature dimension when the batch size of previous samples is 100. In cases where the dimension ratio is low, ``PCR+HC'' performs much better than PCR, and this advantage is maintained as the ratio increases. It means that the feature extraction capability of the model has been improved with the help of HC, allowing for the extraction of more critical features.

\textbf{Temperature component} (HT) is to improve the model's generalization ability. As stated in Table~\ref{ablation}, with the help of HT, the performance of PCR is greatly improved. Compared with ``PCR'' and ``PCR w/o HD'' in Figure~\ref{learningstage5000}, we find that the main improvement brought by HT lies in learning novel knowledge by the model. To explore the role of HT on gradient propagation and model convergence, we analyze the training process of PCR on the CIFAR10 with and without HT, and some important indicators are recorded in Table~\ref{table:ablation_ht}. The results show the final accuracy can be improved by HT since the model can produce less gradient for old classes ($64.28<88.27$) and new classes ($70.40<95.39$) during the gradient propagation process. In the meantime, we find that HT can make the model converge to a region with a lower loss function value ($1.6710<1.8439$) and a flatter landscape during the optimization process. To validate it, we add uniformly distributed noise to the parameters of the final model, causing the position of the model to change in the parameter space. At the same time, we record the differences in training loss before and after the model changes. After repeating the above operations 1000 times, we calculate the average of these 1000 results. The results indicate that the model has relatively small differences using HT for all classes ($0.0025<0.0042$) and new classes ($0.0040<0.0056$). Therefore, the optimal solution of the model using HT is in a relatively flat region. As displayed in Figure~\ref{fig:ablation_ht}, in such a flat region, the model, where the initial solution is $\theta_1$, can obtain a better solution $\theta_2$ than $\theta_3$ for both old and new classes. Moreover, the selection of hyperparameters related to Equation~(\ref{eq:tau_t}) is shown in Tables~\ref{table:tau_step} and \ref{table:tau_bound}. And we set $\tau_{max}=0.16$, $\tau_{min}=0.05$, and $S=500$ based on the results.

\textbf{Distillation component} (HD) is to improve the anti-forgetting ability of the model. Compared with ``HPCR'' and ``HPCR w/o HD'' in Figure~\ref{learningstage5000}, we can find that HPCR performs significantly better on historical knowledge with the help of HD. Shown as Table~\ref{ablation}, the distillation component, which contains SCD and PCD, further improves the performance of PCR. To improve the overall performance of the model, both SCD and PCD are essential. For one thing, SCD can better balance the distribution of historical knowledge while improving the model's ability to anti-forgetting. Actually, there is an imbalance between old classes in OCL. As shown in Table~\ref{table:ablation_hd}, although tasks 1-4 are all old tasks, the performance of the model on these tasks is different. Compared with the distillation method in DER, PCD can better balance the performance of the model on old tasks (Tasks 1-4). For another thing, SCD, which directly propagates gradient for feature extractor, can produce less gradient for all proxies than PCD. As shown in Figure~\ref{fig:ablation_hd}, if HPCR only uses SCD, the gradient is relatively small, and the final performance is better.

In summary, based on these three components, HPCR has a comprehensive improvement compared to PCR. At the same time, all of the components are tailored to break the limitations of PCR and have their own originality.

\subsection{Further study}

\subsubsection{Pre-trained Backbone} With the development of pre-trained models, continual learning tends to take a pre-trained model as the initial model. Hence, we further conduct experiments with a pre-trained model on Split CIFAR100. For fairness, we consider the ViT-B/16-IN21K in~\cite{zhou2024expandable} as the pre-trained model, where each frozen transformer block contains a trainable residual term. Meanwhile, the training batch size is set as 10 and the epoch is set as 1 for all methods. As stated in Figure~\ref{fig:pretrained}, SLCA~\cite{zhang2023slca} and EASE~\cite{zhou2024expandable} are non-replay methods while ER replays with a 100-size buffer. Compared with ER (100) and HPCR (100), we can find that HPCR can also play a role when using a pre-trained backbone. Moreover, the performance of HPCR will gradually surpass SLCA and EASE based on pre-trained models as the buffer size increases (from 100 to 5000), further confirming its superiority for OCL.

\subsubsection{Visualization} Visualization can
help reflect the role of these components more intuitively. We train the model by ER and HPCR on Split CIFAR10 with 500 sizes of memory buffer, respectively. At the end of the training, we report their 2D t-SNE~\cite{van2008visualizing} visualization of feature embeddings for all testing samples, as shown in Figure~\ref{fig:tsne}. The stars are proxies while others are samples for all classes. The results demonstrate that HPCR not only achieves great performance, but also can indeed improve the distinguishability of samples in the embedding space.

\subsubsection{Computation and Memory Cost} For OCL, the overhead of memory and computing resources is crucial, since it will affect the practicality of a method. Therefore, we analyze the memory and computational cost of different methods, as reported in Table~\ref{table:other}. 1) Similar to~\cite{ghunaim2023real}, we use $C_\mathcal{D}$ to compare the computation of different methods. The $C_\mathcal{D}$ is denoted as a relative complexity between the stream and an underlying continual learning method. For example, when current samples come, ER can immediately learn them and then predict unknown samples, resulting in a relative complexity of 1. Since UER, ER-ACE, PCR, and HPCR only modify the loss objective of ER, their computational complexities are equivalent to 1. However, OCM and OnPro require an additional 16 augmentation to the original data, making it require 16  times the FLOPs needed by ER. Thus, its computational complexity is 16, which limits the ability of real-time prediction. 2) We set the memory budget of the model for ER as 1. Due to the OCM method requiring an additional model to be saved for knowledge distillation, its relative budget is 2 and others are 1. 3) We evaluate the memory usage of different methods by considering the data size of all samples stored in the buffer. As seen in Table~\ref{table:memory}, since HPCR has additional storage requirements for features and logits embedding, it necessitates relatively more memory space ($>1$). However, the additional memory does not occupy a high proportion ($<0.1$). If the additional memory is used to store old samples, the performance of the model is shown in the row of ``Accuracy (Control)''. The results demonstrate that storing features and logits as HPCR yields significant benefits, justifying its worth. In summary, HPCR emerges as a straightforward yet efficacious approach for real-time prediction in OCL scenarios.

\section{Conclusion}
\label{sec:conclusion}
In this paper, we develop a more holistic OCL method named HPCR based on PCR, which mainly consists of three components. The contrastive component conditionally introduces anchor-to-sample pairs to PCR, improving the feature extraction ability of PCR when the training batch size is large. The temperature component decouples the influence of temperature on gradients in two parts and effectively improves the generalization ability of the model. The distillation component improves the anti-forgetting ability of the model through PCD and SCD. Extensive experiments on four datasets demonstrate the superiority of HPCR over various state-of-the-art methods.


 
%

\bibliographystyle{IEEEtran}

\begin{IEEEbiography}[{\includegraphics[width=1in,height=1.25in,clip,keepaspectratio]{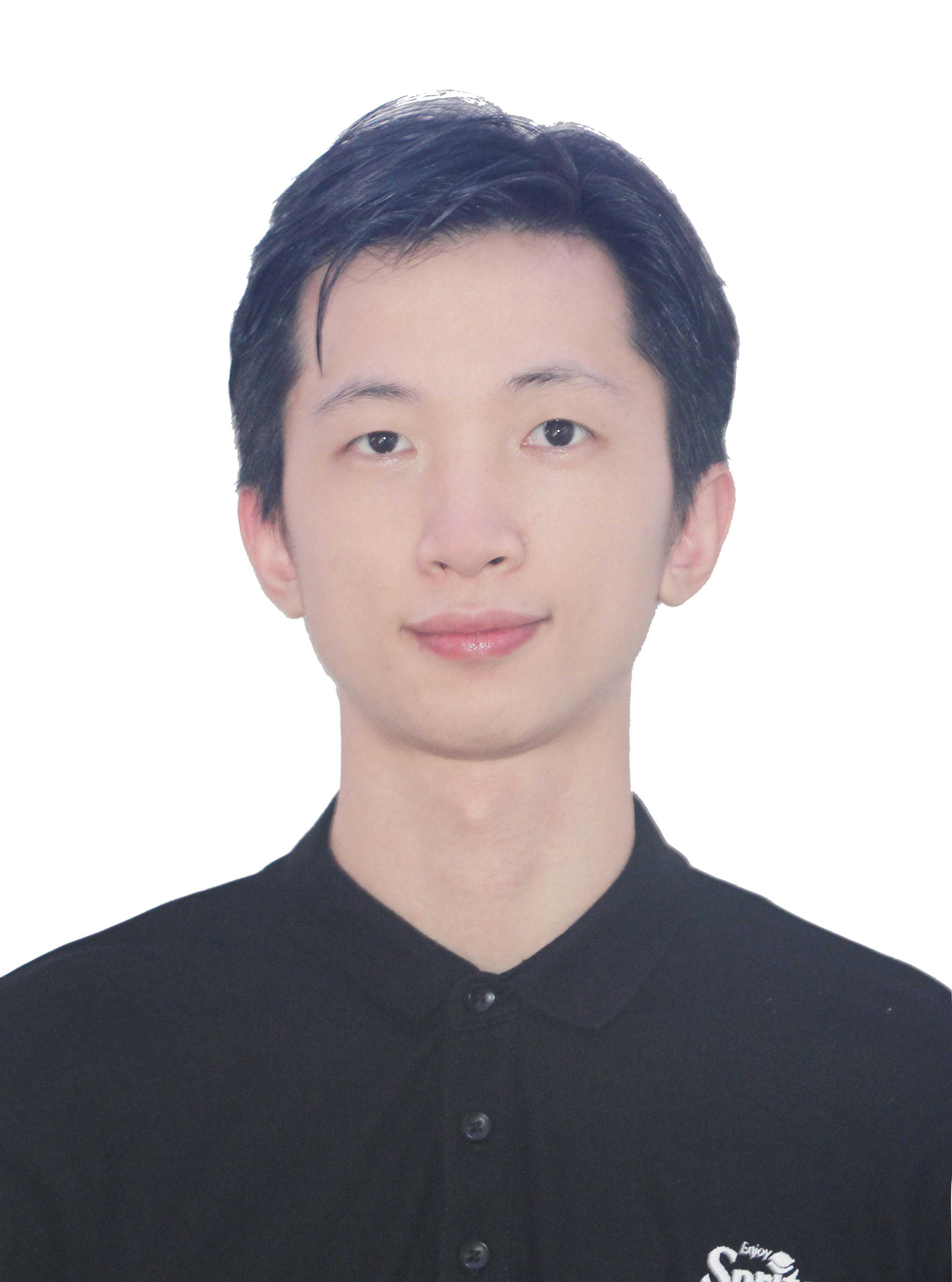}}]{Huiwei Lin} is is currently a Postdoctoral Fellow at The Chinese University of Hong Kong. He received the B.S. degree from the South China University of Technology, China, in 2017, the M.S. degree from the Harbin Institute of Technology, Shenzhen, China, in 2020, and the Ph.D. degree from the Harbin Institute of Technology, Shenzhen, China, in 2024. His current research interests include continual learning and knowledge reasoning.
\end{IEEEbiography}
\vspace{-5ex}
\begin{IEEEbiography}[{\includegraphics[width=1in,height=1.25in,clip,keepaspectratio]{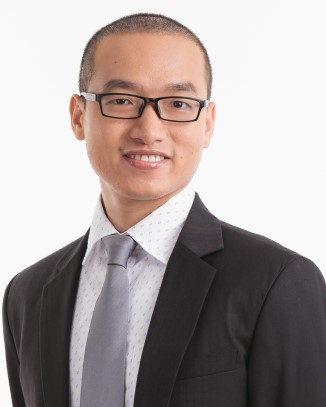}}]{Shanshan Feng} is currently a senior research scientist at the Centre for Frontier AI Research, Institute of High Performance Computing, the Agency for Science, Technology and Research (A*STAR), Singapore. He received the Ph.D. degree from Nanyang Technological University, Singapore in 2017, and his B.S. degree from University of Science and Technology of China in 2011. His current research interests include spatial-temporal data mining, social graph learning, and recommender systems. 
\end{IEEEbiography}
\vspace{-5ex}
\begin{IEEEbiography}[{\includegraphics[width=1in,height=1.25in,clip,keepaspectratio]{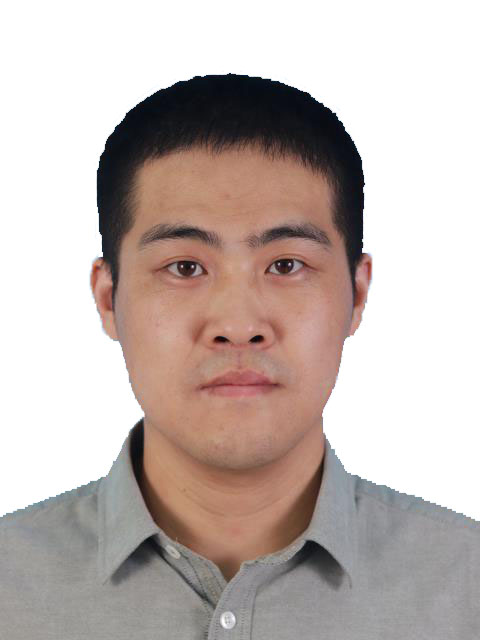}}]{Baoquan Zhang} is currently an assistant professor with the School of Computer Science and Technology, Harbin Institute of Technology, Shenzhen. He received the B.S. degree from the Harbin Institute of Technology, Weihai, China, in 2015, the M.S. degree from the Harbin Institute of Technology, Harbin, China, in 2017, and the Ph.D. degree from the Harbin Institute of Technology, Shenzhen, in 2023. His current research interests include meta learning, few-shot learning, machine learning, and data mining.
\end{IEEEbiography}
\vspace{-5ex}
\begin{IEEEbiography}[{\includegraphics[width=1in,height=1.25in,clip,keepaspectratio]{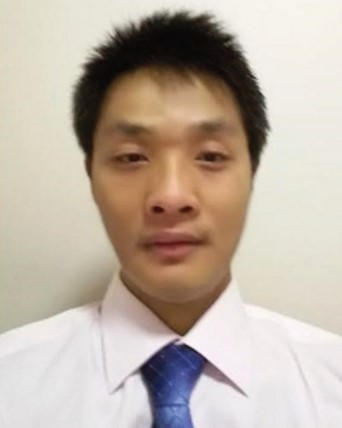}}]{Xutao Li} is currently a Professor with the School of Computer Science and Technology, Harbin Institute of Technology, Shenzhen, China. He received the Ph.D. and Master degrees in Computer Science from Harbin Institute of Technology in 2013 and 2009, and the Bachelor from Lanzhou University of Technology in 2007. His research interests include data mining, machine learning, graph mining, and social network analysis, especially tensor-based learning, and mining algorithms.
\end{IEEEbiography}
\vspace{-5ex}
\begin{IEEEbiography}[{\includegraphics[width=1in,height=1.25in,clip,keepaspectratio]{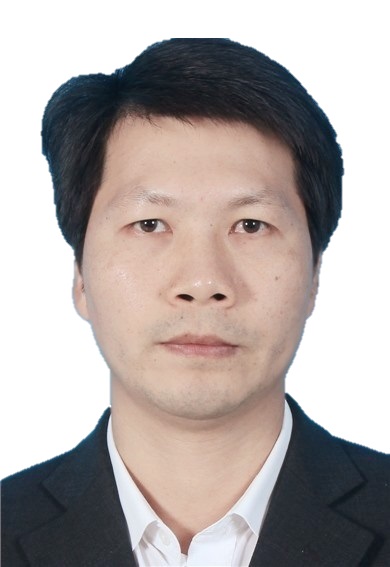}}]{Yunming Ye} is currently a Professor with the School of Computer Science and Technology, Harbin Institute of Technology, Shenzhen, China. He received the PhD degree in Computer Science from Shanghai Jiao Tong University, Shanghai, China, in 2004. His research interests include data mining, text mining, and ensemble learning algorithms.
\end{IEEEbiography}

\appendix[Proof of Theorem 1]
By chain-rule, we have

 \begin{equation}
    \frac{\partial L_{PCR}}{\partial <\bm{z},\bm{w}_c>}=(\frac{\partial L}{\partial p^*_c}\frac{\partial p^*_c}{\partial f(\bm{z};\bm{W})})\frac{\partial f(\bm{z};\bm{W})}{\partial <\bm{z},\bm{w}_c>}.
\end{equation}
If $c=y$, we can get

\begin{equation}
    \begin{split}
    &
    \frac{\partial L_{PCR}}{\partial p^*_c}\frac{\partial p^*_c}{\partial f(\bm{z};\bm{W})} \\
    &
    =\frac{-1}{p^*_y}\frac{exp(o_y)(\sum_{c\in\mathcal{C}_{1:t}}exp(o_c)-k_y exp(o_y))}{(\sum_{c\in\mathcal{C}_{1:t}}k_cexp(o_c))^2}
    \\
    &
    =\frac{-1}{p^*_y}(p^*_y-k_yp^*_yp^*_y)
    =k_yp^*_y-1.
    \end{split}
\end{equation}
Otherwise, if $c\neq y$, we can get
\begin{equation}
    \begin{split}
    &
    \frac{\partial L_{PCR}}{\partial p^*_c}\frac{\partial p^*_c}{\partial f(\bm{z};\bm{W})} \\
    &
    =\frac{-exp(o_c)}{p^*_cexp(o_y)}\frac{-k_cexp(o_c)exp(o_y)}{(\sum_{c\in\mathcal{C}_{1:t}}k_cexp(o_c))^2}
    \\
    &
    =\frac{-1}{p^*_c}(-k_cp^*_cp^*_c)
    =k_cp^*_c.
    \end{split}
\end{equation}
Since $\partial f(\bm{z};\bm{W})/\partial <\bm{z},\bm{w}_c>=1/\tau$, the final gradient can be denoted as 

 \begin{equation}
    \label{eq:gradient_pcr_a}
    \frac{\partial L_{PCR}}{\partial <\bm{z},\bm{w}_c>}=\left\{
    \begin{aligned}
        (k_yp^*_y-1)/\tau&,&c= y \\
        (k_cp^*_c)/\tau&,&c\neq y
    \end{aligned}
    \right..
\end{equation}

\vfill

\end{document}